%% file: icml2024.tex
\documentclass{article}
\usepackage{microtype}
\usepackage{graphicx}
\usepackage{subcaption}
\usepackage{array}
\usepackage{multirow, makecell}
\usepackage{booktabs} % for professional tables

\usepackage{hyperref}
\usepackage[dvipsnames]{xcolor}

 \usepackage{wrapfig}
 \newcolumntype{C}[1]{>{\centering\arraybackslash}m{#1}}
\usepackage[accepted]{icml2024}

\usepackage{amsmath}
\usepackage{amssymb}
\usepackage{mathtools}
\usepackage{amsthm}
\newcolumntype{P}[1]{>{\centering\arraybackslash}p{#1}}

\usepackage[capitalize,noabbrev]{cleveref}

\theoremstyle{plain}

\theoremstyle{definition}

\theoremstyle{remark}

\usepackage[textsize=tiny]{todonotes}

\icmltitlerunning{Structure Your Data: Towards Semantic Graph Counterfactuals}

\begin{document}

\twocolumn[
\icmltitle{Structure Your Data: Towards Semantic Graph Counterfactuals}



\icmlsetsymbol{equal}{*}

\begin{icmlauthorlist}
\icmlauthor{Angeliki Dimitriou}{yyy}
\icmlauthor{Maria Lymperaiou}{yyy}
\icmlauthor{Giorgos Filandrianos}{yyy}
\icmlauthor{Konstantinos Thomas}{yyy}
\icmlauthor{Giorgos Stamou}{yyy}

\end{icmlauthorlist}

\icmlaffiliation{yyy}{Artificial Intelligence and Learning Systems Laboratory, National Technical University of Athens}

\icmlcorrespondingauthor{Angeliki Dimitriou}{angelikidim@ails.ece.ntua.gr}
%\icmlcorrespondingauthor{Firstnameangelikidim@ails.ece.ntua.gr2 Lastname2}{first2.last2@www.uk}

\icmlkeywords{Counterfactual Explanations, Graph Neural Networks, Scene Graphs, Explainable Artificial Intelligence
}

\vskip 0.3in
]



\printAffiliationsAndNotice{} % otherwise use the standard text.

\begin{abstract}
Counterfactual explanations (CEs) based on concepts are explanations that consider alternative scenarios to understand which high-level semantic features contributed to particular model predictions.
In this work, we propose CEs based on the semantic graphs accompanying input data to achieve more descriptive, accurate, and human-aligned explanations. 
Building upon state-of-the-art (SotA) conceptual attempts, we adopt a model-agnostic edit-based approach and introduce leveraging GNNs for efficient Graph Edit Distance (GED) computation. 
With a focus on the visual domain, we represent images as scene graphs and obtain their GNN embeddings to bypass solving the NP-hard graph similarity problem for all input pairs, an integral part of CE computation process. 
We apply our method to benchmark and real-world datasets with varying difficulty and availability of semantic annotations. Testing on diverse classifiers, we find that our CEs outperform previous SotA explanation models based on semantics, including both white and black-box as well as conceptual and pixel-level approaches. Their superiority is proven quantitatively and qualitatively, as validated by human subjects, highlighting the significance of leveraging semantic edges in the presence of intricate relationships. 
Our model-agnostic graph-based approach is widely applicable and easily extensible, producing actionable explanations across different contexts. The code is available at \url{https://github.com/aggeliki-dimitriou/SGCE}.
\end{abstract}

\input{sections/introduction}
\input{sections/related}
\input{sections/method}

\input{sections/experiments}

\input{sections/conclusion}
\input{sections/impact}

\bibliography{icml2024}
\bibliographystyle{icml2024}

%%%%%%%%%%%%%%%%%%%%%%%%%%%%%%%%%%%%%%%%%%%%%%%%%%%%%%%%%%%%%%%%%%%%%%%%%%%%%%%
%%%%%%%%%%%%%%%%%%%%%%%%%%%%%%%%%%%%%%%%%%%%%%%%%%%%%%%%%%%%%%%%%%%%%%%%%%%%%%%
% APPENDIX
%%%%%%%%%%%%%%%%%%%%%%%%%%%%%%%%%%%%%%%%%%%%%%%%%%%%%%%%%%%%%%%%%%%%%%%%%%%%%%%
%%%%%%%%%%%%%%%%%%%%%%%%%%%%%%%%%%%%%%%%%%%%%%%%%%%%%%%%%%%%%%%%%%%%%%%%%%%%%%%
\newpage
\appendix
\onecolumn
%\section{Appendix}

\input{sections/appendix}

%%%%%%%%%%%%%%%%%%%%%%%%%%%%%%%%%%%%%%%%%%%%%%%%%%%%%%%%%%%%%%%%%%%%%%%%%%%%%%%
%%%%%%%%%%%%%%%%%%%%%%%%%%%%%%%%%%%%%%%%%%%%%%%%%%%%%%%%%%%%%%%%%%%%%%%%%%%%%%%

\end{document}

%% file: sections/introduction.tex
\section{Introduction}
\label{sec:intro}
    \begin{figure}[t]
    % \vskip 0.2in
        \centering
% \vspace{-0.5cm}
% \hskip -0.05in
\includegraphics[width=1.\columnwidth]{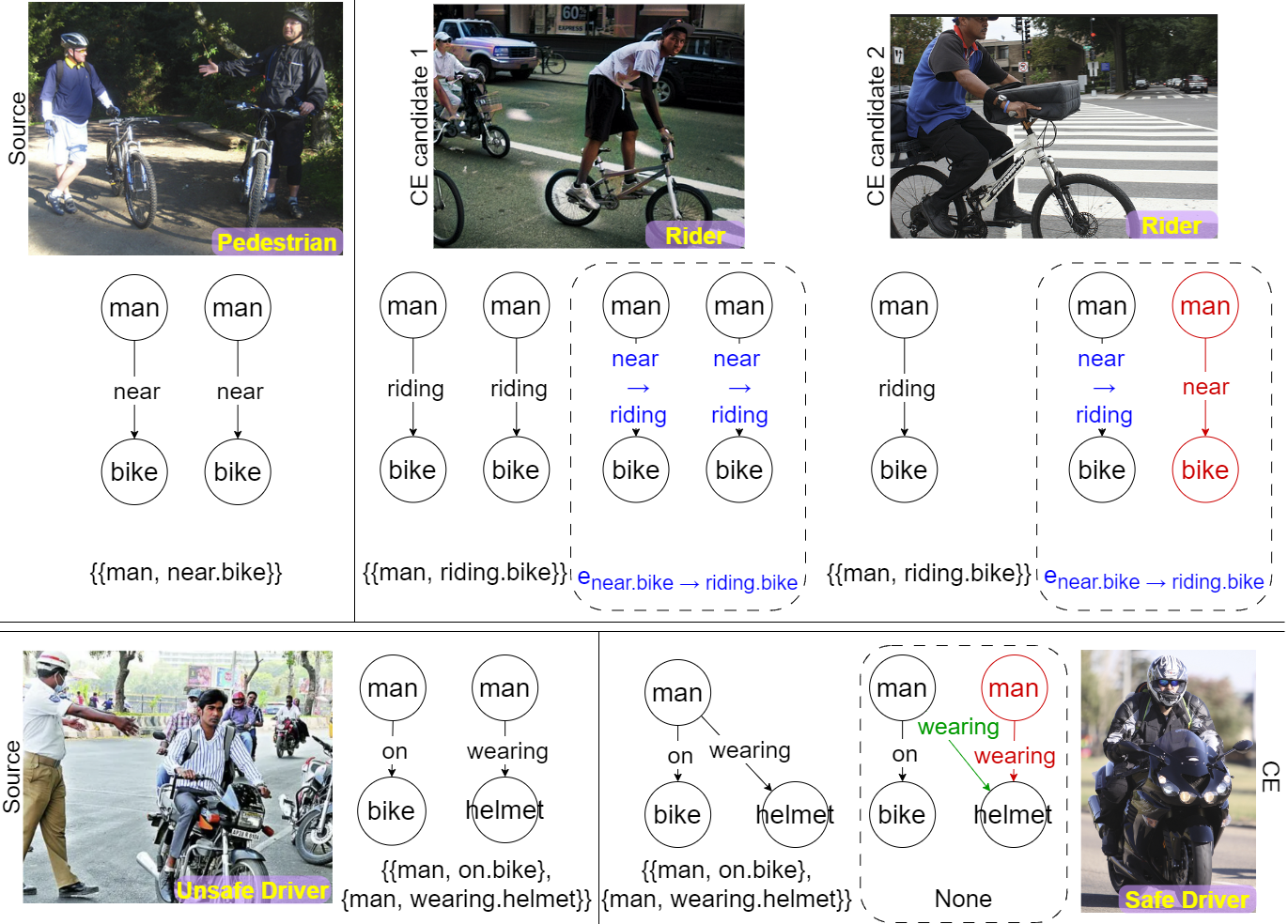}
% \vspace{-0.5cm}

\caption{Examples where semantic graphs trump concept sets. Example 1 (top) shows the importance of the multiplicity of concepts for edit distance and example 2 (bottom) emphasizes the intricacy of relations. Edits (\textcolor{blue}{substitutions}, \textcolor{ForestGreen}{insertions}, \textcolor{red}{deletions}) are enclosed in striped rectangles. Images sourced from Visual Genome \cite{krishna2017visual}, except unsafe driver \cite{Deccan}.}
% \vskip 0.1in
        \label{fig:example1}
        \vskip -0.15in
    \end{figure}

The AI landscape, now dominated by advanced Large Multimodal Models such as GPT4, GPT4V, and Gemini, highlights the widespread use of proprietary models due to their state-of-the-art (SotA) performance across various modalities and datasets\footnote{https://openai.com/research/image-gpt} \citep{gemini}. This underscores the need for increased attention to black-box explainability methods, especially with the growing related applications in critical areas like medical image classification \citep{wu2023can, gpt4v}. Users should have the ability to understand decision-making processes without accessing the classifier architecture, emphasizing the importance of autonomy in scrutinizing proprietary models. To address this, there is a rising demand for post-hoc/model-agnostic explainability, an established field with publications in prestigious conferences \citep{ribeiro2016should, ying2019gnnexplainer, wisely}. In this spirit, this paper proposes a black-box method to compute Counterfactual Explanations (CEs) \citep{wachter2017counterfactual} based on semantics. The challenges posed to extract and interpret decision processes of black-box models, although acknowledged as inherent trade-offs \citep{rudin2019stop}, lie beyond the scope of this work.
    
The effectiveness of Conceptual XAI methods is closely tied to the semantic context of the data they interpret. In fact, \citet{browne2020semantics} report that 'there is no explanation without semantics', and formally prove that semantics are the distinguishing factor between CEs and adversarial examples. The role of annotations as an integral component in formulating conceptual CEs was first highlighted in \citet{filandrianos2022conceptual}%(CECE)
, where the term "Explanation Dataset" was introduced. Its curation is also the initial step of the counterfactual computation pipeline proposed by the SotA CE work of \citet{wisely} (SC), which emphasizes its significance by urging users to 'choose their data wisely'. Given that the ultimate recipients of the explanations are humans, it is crucial to select annotations with precision and to actively engage domain experts in the process. This ensures that the explanations are not only accurate but also meaningful and relevant to the intended audience.

In the context of visual CEs, leveraging semantic annotations instead of superficial pixel-based features is significant, but not sufficient if their relations are not represented accurately. Our work addresses such limitations by \textit{structuring the semantics as a graph}. In Fig. \ref{fig:example1}, we see a coarse representation of depicted concepts and their relations for source images and CE candidates, using our proposed semantic graphs versus the set of sets representation of SC. The explanations we provide include counterfactual images accompanied by the edit graphs from source to counterfactual. Fig. \ref{fig:example1} (top) illustrates an example where employing sets under-represents the edit number by treating the two CE candidates as equals, despite the varying number of pedestrians (man riding bike) between them, potentially leading to a CE that is not optimal neither in terms of Graph Edit Distance (GED), nor visually. In other words, SC would consider a picture of a single rider the same as a photo of tens of cyclists. Fig \ref{fig:example1} (bottom) depicts another problematic case for SC, where for the same CE the edit path is misleading. Using set representation, it is unclear that the rider lacks a helmet in the source image, creating the false sense of no required semantic edits. In contrast, our graph method recommends adding the 'wearing' role between 'man on bike' and 'helmet'. For 'safe' vs 'unsafe' driving classification, this edit path is crucial for explanations, both locally and globally. These two instances motivate the expressivity of graph-based explanations. By further linking semantics with external knowledge, we constrain edits to establish that concepts such as 'man' and 'woman' are more closely related than concepts like 'man' and 'helmet'; thus, boosting the interpretability and actionability of our method.  
    
This work serves as an advancement of the previously presented method by \citet{wisely}, which emphasizes leveraging semantically rich concepts to obtain CEs within model-agnostic settings, as long as the participating data instances are chosen wisely. However, their data representation lacks the proper incorporation of relationships between concepts, calling for a more intricate approach. We not only employ graphs to structure semantic information as a direct refinement of prior work, but also leverage Graph Neural Networks (GNNs) for the efficient approximation of GED between graph instances to compute CEs. Our findings confirm the significance of correctly representing the number and interactions of concepts and
% demonstrate that the richer and more expressive graph structure offers significant benefits in scenarios where the number of objects and the relationships between them are crucial in distinguishing between two classes. 
% In addition, 
our method 
significantly narrows the gap to the golden standard GED, achieving closer proximity with fewer edits. To underscore the efficacy of our approach against methods with access to the underlying model, we expanded the human survey from SC and compared our CEs with the white-box CE method by \citet{vandenhende2022making} (CVE), which, despite being pixel-level, emphasizes preserving semantic consistency. Our evaluation aimed to assess both human preferences and their capacity to comprehend and anticipate the classifier's output.

Our survey revealed that participants preferred our CEs in the majority of instances, and they were also successful in learning to accurately classify images themselves. This indicates that our approach surpasses the explanatory power of the white-box method of CVE in clarifying classifier logic to humans. This finding was further reinforced by replicating this experiment \textit{exclusively providing semantic graphs and edits to the users, without any images}. Participants comprehended the classifier's reasoning and predicted outcomes effectively, even without the corresponding visual data.

We prove that our work surpasses prior SotA approaches. Our improvements have been quantitatively assessed and further substantiated by human surveys - a significant XAI evaluation tool. The compared methodologies are diverse in terms of reliance on the features %of the machine learning model 
classifiers exploit and the granularity of information. To combat the challenges of evaluating the explanations, we establish unified quantitative and qualitative metrics, applicable in all cases. Further elaboration is available in the Evaluation section of \S \ref{sec:experiments}. Our key contributions are: 
\begin{itemize}
    \item Demonstrating  \textbf{quantitative and qualitative superiority} over previous black- and white-box methods, paired with \textbf{enhanced adaptability} due to our model-agnostic design, 
    \item Offering more \textbf{interpretable, expressive, and actionable} CEs using semantic graphs, 
    \item Achieving \textbf{efficient GED approximation}  via GNNs without compromising the representation of concept interactions.
\end{itemize}

Our method is \textbf{novel} in employing graphs and GNNs for counterfactual retrieval. We validate our approach across four diverse datasets (images \& audio), using three neural and one non-neural classifier, including two human surveys and four quantitative and qualitative experiments, yielding superior outcomes and time efficiency relative to SotA.

%% file: sections/related.tex
\section{Related Work}
\label{sec:related}

\begin{figure*}[!t]
% \vskip 0.2in
  %\hspace{-7px}
  \centering
  % \vskip -0.01in
  \includegraphics[width=0.95\textwidth]{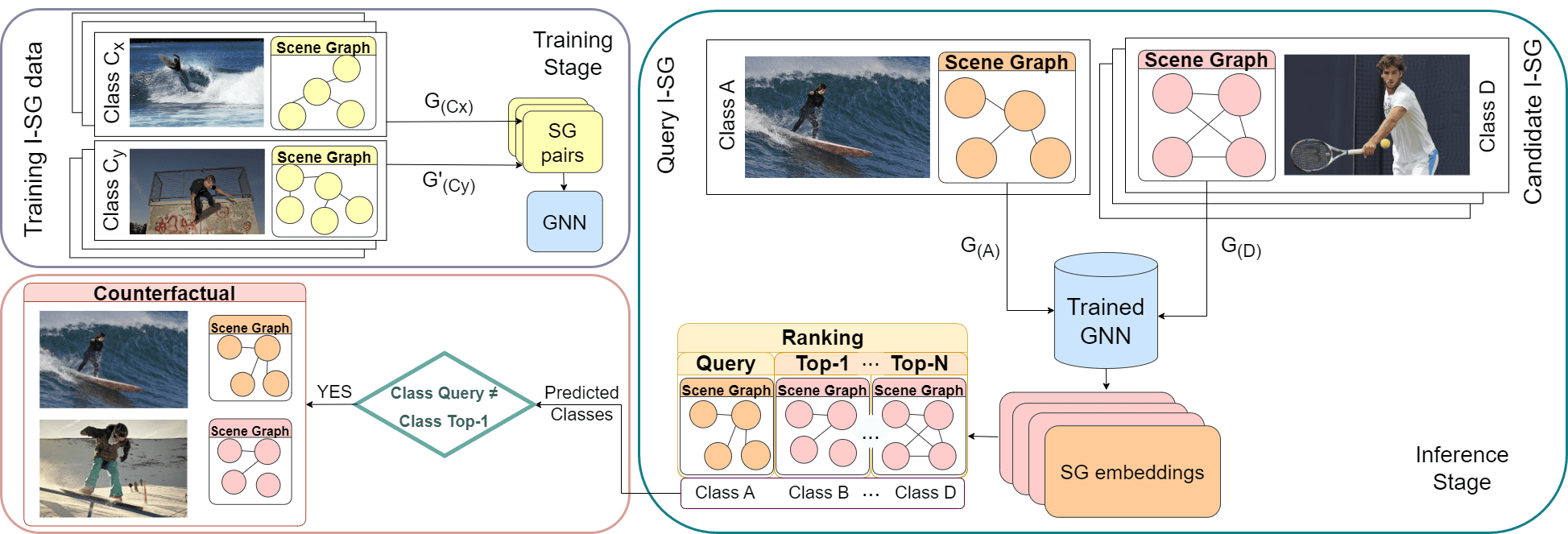}
    %\vskip 0.1in
  \caption{Method outline (for image classifiers). Depicted stages directly correspond to Sec. \ref{sec:method} paragraphs. \textit{Predicted} class labels are: A - query, B - target, $C_x$, $C_y$ - any class, others - random class instances. Graph $G'_{(B)}$ corresponds to counterfactual image $I'_{(B)}$.}
\label{fig:outline}
% \vskip -0.2in
\end{figure*}

\paragraph{Counterfactual explanations} of visual classifiers encompass pixel-level edit methods which focus on marking and altering significant image areas that influence the model's predictions \citep{pmlr-v97-goyal19a, vandenhende2022making, augustin2022diffusion}. %some even leveraging advanced generative techniques. 
Contrary to other feature extraction counterfactual methods, the Counterfactual Visual Explanations (CVE) of \citet{vandenhende2022making} attempt to enforce semantically consistent area exchanges through an auxiliary semantic similarity component between local regions. Their semantically driven approach is the prime choice for pixel-level comparison, also providing a benchmark in contrast to techniques with direct classifier access, a feature distinguishing it from ours. 
Another vein of research focuses on human-interpretable concept edits to retrieve CEs.
% \citep{filandrianos2022conceptual,  abid2022meaningfully, wisely}. 
\citet{abid2022meaningfully} propose conceptual CEs in the event of a misclassification, using a white-box technique based on Concept Activation Vectors.
The work by \citet{wisely} (SC) serves as an extension of \citet{filandrianos2022conceptual}, additionally emphasizing the importance of the Explanation Dataset, and expanding upon the original graph bipartite matching CE framework by leveraging the roles between concepts. Our work directly enhances these approaches, retaining advantageous qualities such as their model-agnostic nature and definition of object/relation distance through ontologies. Instead of leveraging Set Edit Distance %as a proxy 
and ignoring edges %of the semantic graph 
altogether \citep{filandrianos2022conceptual} or rolling up the edges into concepts, thus sacrificing crucial object relation information \citep{wisely}, we use the more accurate GED. To this end, we introduce semantic graphs for increased expressivity and use GNNs to accelerate GED calculation. A preliminary GNN-based counterfactual analysis was thoroughly discussed in the concurrent work of \cite{dimitriou2024graph}, which compares various Graph Machine Learning algorithms.
Much like its predecessors, our current paper is centered on the visual domain and applied to other modalities as a use case. Consequently, we do not delve into the literature of audio CEs, for instance.
Despite the vast literature on graph CEs \cite {prado2024survey}, comparison to GNN explainers \citep{NEURIPS2021_2c8c3a57,lucic2022cf} is not applicable here, since we propose utilizing semantic graphs corresponding to non-graph input data for CE computation. It is noteworthy that no existing method leverages GNNs for post-hoc CEs, a novel feature of our approach.

\paragraph{Graph similarity} methods such as Graph Edit Distance (GED) \citep{sanfeliu1983distance} are computationally expensive, prompting the use of approximation algorithms.
% we say it in method, we just use it to compute ground truth we don't build upon it
% \citet{fankhauser2011speeding} introduced a GED approximation that combines the Jonker-Volgenant assignment algorithm with a bipartite heuristic leading to significant speedup, which we adopt for the fast computation of GED during training. 
Considering neural approaches, the ones relevant to our work leverage GNNs 
\citep{bai2019unsupervised, li2019graph, ranjan2022greed}. %could remove
% These techniques commonly involve training two identical GNNs using graph pairs as input and their similarity to compute loss. 
As our paper focuses on embedding extraction to facilitate CEs instead of the similarity itself, we simply draw inspiration from previous approaches in implementing our GNN model for semantic graphs, by adopting the ideas of Siamese GNNs, graph-to-graph proximity training and Multi-Dimensional Scaling as loss \citep{bai2019unsupervised} to preserve inter-graph distances in the embedding space.

%% file: sections/method.tex
\section{Method}
\label{sec:method}

Since the majority of our experiments are conducted with visual classifiers, we will illustrate our framework within this domain (Fig. \ref{fig:outline}). Given a query image $I_{(A)}$ belonging to a class $A$, a conceptual CE entails finding another image $I_{(B)}' \neq I_{(A)}$ in a class $B \neq A$, so that the shortest edit path between $I_{(A)}$ and $I_{(B)}'$ is minimized. Even though there are different notions of distance between images, we select a \textit{conceptual} representation, employing scene graphs to represent objects and interactions within images. To this end, the problem of image similarity ultimately reduces to a graph similarity challenge.
However, graph edits (insertions, deletions, substitutions) as a deterministic measure of similarity between two graphs $G_{(A)}$ and $G_{(B)}'$ is an NP-hard problem. 
Optimal edit paths can be found through tree search algorithms with the requirement of exponential time.
When searching for a counterfactual graph to $G_{(A)}$ among a set of $N$ graphs, GED needs to be calculated $N-1$ times.
To minimize the computational burden, we use lightweight GNNs that accelerate the graph proximity process by mapping all $N$ graphs to the same embedding space.
By retrieving the closest embedding to $G_{(A)}$ that belongs to class $B \neq A$, GED is computed only \textit{once} per query during retrieval. Concretely, we approximate the following optimization problem for semantic graphs extracted from any input modality:
\begin{equation}
    \textit{GED}(min|G_{(A)}, G_{(B)}'|), \textit{ such that } A \neq B
\end{equation}

% We note that the following procedure could be appropriately adapted to any modality beyond images and their corresponding scene graphs (App.  \ref{}). 

\paragraph{Ground Truth Construction}
As our overall approach does not rely on pre-annotated graph distances, we propose a technique to construct well-defined ground truth instances. The graph structure of data imposes the requirement of defining an absolute similarity metric between graph pairs for the training stage. GED is regarded as the optimal choice despite its computational complexity; computing GED for only $N/2$ pairs to construct the training set is adequate for achieving high quality representations, as validated experimentally. To further facilitate 
%ground truth 
GED calculation, we exploit a suboptimal algorithm utilizing a bipartite heuristic that accelerates an already effective in practice LSAP-based algorithm for GED \citep{jonker1987shortest, fankhauser2011speeding}.
Consequently, semantic information of nodes and edges should guide graph edits based on their conceptual similarity. Thus, we choose to deploy the technique proposed in SC \citep{wisely} to
assign operation costs based on conceptual edit distance, as instructed by the shortest path between two concepts within the WordNet hierarchy \citep{miller1995wordnet}. 

\paragraph{GNN Training}
To accelerate the retrieval of the most similar graph $G_{(B)}'$ to graph $G_{(A)}$, we build a siamese GNN component for graph embedding extraction based on inter-graph proximity. The GNN comprises two identical node embedding units that receive a random graph pair $(G_{(C_x)}, G_{(C_y)}')$ as input  ($C_x, C_y$ can be any class). The extracted node representations are pooled to produce global graph embeddings $(h_{G_{(C_x)}}, h_{G_{(C_y)}'})$. Embedding units consist of stacked GNN layers, described by either GCN \citep{kipf2016semi}, GAT \citep{velivckovic2017graph} or GIN \citep{ xu2018powerful}. We formalize GCN graph embedding computation in Eq. \ref{eq:gcn_embed} (omitting class notation for simplicity):
\begin{equation}
    h_{G} = \frac{1}{n} \sum_{i=1}^{n}(u_i^{K-1} + \sum_{j \in \mathcal{N}(i)}u_j^{K-1})
    \label{eq:gcn_embed}
\end{equation}
where $u_i$ is the representation of node $i$, $\mathcal{N}(i)$ is the neighborhood of $i$, $n$ is the number of nodes for $G$ and $K$ is the number of GCN layers. 
To preserve the similarity of vectors $(h_{G_{(C_x)}}, h_{G_{(C_y)}'})$, we adopt the dimensionality reduction technique of Multi-Dimensional Scaling \citep{williams2000connection}, as proposed in \cite{bai2019unsupervised}. The model is trained transductively to minimize the loss function $\mathcal{L}$:
\begin{equation}
   \mathcal{L} = \mathbb{E}(\left \| (h_{G_{(C_x)}} - h_{G_{(C_y)}'} \right \|_2^2 - GED(G_{(C_x)}, G_{(C_y)}'))
\end{equation}
Graphs are 
%successfully 
embedded in a lower dimensional space by choosing a random subset of $\frac{N!}{2(N-2)!}$ pairs with varying cardinality $p$. %, where $N$ is the number of graphs in the dataset. 
As node features initialization is significant with regard to semantic similarity preservation, we use GloVe representations \citep{pennington2014glove} of 
%the 
node labels.

\paragraph{Ranking and Counterfactual Retrieval}
Once graph embeddings have been extracted, they are compared using cosine similarity to produce rankings. For each query image $I_{(A)}$ and subsequently its scene graph $G_{(A)}$, we obtain the instance $G_{(B)}'$ with the highest rank given the constraint that $I_{(B)}'$ is classified in $B \neq A$. $I_{(B)}'$ is proposed as a CE of $I_{(A)}$ since it constitutes the instance with the minimum graph edit path from it, classified in a different target category $B$. Specifically, we retrieve a scene graph $G_{(B)}'$ as:
\begin{equation}
    G_{(B)}' = G_{(B)}^i \textit{, }\arg \max_{i}(\frac{h_{G_{(B)}^i}\cdot h_{G_{(A)}}}{\left \| h_{G_{(B)}^i} \right \|\left \| h_{G_{(A)}} \right \|}) \textit{ if } B \neq A  
\end{equation}
where  $i = 1, ..., N$. Selecting target class $B$ is %strongly 
correlated with the characteristics of the dataset in use and the goal of the explanation itself. Precisely, if the data instances have ground truth labels, the target class can be defined as the most commonly confused compared to the source image class \citep{vandenhende2022making}. Another valid choice is to arbitrarily pick $B$ to facilitate a particular application, i.e. explanation of classifier mistakes, in which case $B$ is the true class of the query image \cite{abid2022meaningfully}. We choose the first approach when ground truth class labels are available; otherwise, we define the target class as the one with the most highly ranked instance not classified as A.
% $c' \neq c$, 
% differently than source image $I_{(A)}$.

%% file: sections/experiments.tex
\section{Experiments}
\label{sec:experiments}

%\paragraph{Experimental settings}

\paragraph{Evaluation} comprises quantitative metrics, as well as human-in-the-loop experiments.
Quantitative results are extracted by comparing the ranks retrieved based on our obtained graph embeddings to the ground truth ranks retrieved by GED. 
This type of analysis is not present for SC, despite its significance for objectively assessing CEs beyond intuitive metrics.
The reported metrics are: 1) \textit{average Precision@k (P@k)}: all top-k GED retrieved results are considered relevant, 2) \textit{binary P@k} and \textit{binary NDCG@k}: only top-1
GED result is relevant and its position in retrieved ranks is emphasized through NDCG, 3) \textit{average number of edits}: average number of node/edge insertions, deletions, and substitutions with different concepts, calculated post-hoc through GED to ensure fairness.

The use of GED rankings as the golden standard for evaluation is clearly motivated by previous work (SC) and reinforced by features like: a) its purely semantic nature, b) completeness in distance representation due to its reliance on graphs which accurately encompass both objects and relations, c) deterministic nature and applicability regardless of modality and granularity of the technique under evaluation. Wide applicability is especially important because baselines include pixel-based methods which define units of information differently (significant rectangular areas vs concepts). Thus, investigating the effectiveness of GED in depth or comparing it with other similar metrics would divert from the paper's main focus, as it is already accepted in the research community. 
% Of course, all this requires that "data was chosen wisely" beforehand.    

Human evaluation highlights several aspects of our contributions. 
First, to validate the quality of our retrieved CEs against SotA, we ask our evaluators (student volunteers of engineering backgrounds) to select among two CE alternatives of a query image; an image retrieved from our method versus an image retrieved either by SC or CVE. 
We also test the understandability of our CEs by replicating the machine-teaching human experiment of CVE, adjusted to accommodate our graph-based explanations. We design the same stages (pre-learning, learning, and testing) and equally divide our annotators into two independent learning stage variants, namely 'visually-informed' and 'blind'. The 'blind' variant is the only different setting from CVE: the annotators of the 'blind' learning stage are only provided with scene graph pairs and graph edits but no images. This evaluation method is being used for the first time to measure the reliance of humans on graph concepts rather than visual cues to understand the reasoning for classification. %Examples of graphs and corresponding edits are provided in App.  \ref{sec-app:graphs}.
More information about human evaluation is provided in App. \ref{sec:human}.

\paragraph{Experimental settings and objectives} Our presented results involve $p\sim N/2$ training graph pairs and the GCN variant unless mentioned otherwise. We produce graph representations using a single Tesla K80 GPU, while all other computations are done on a 12-core Intel Core i7-5930K CPU. We utilize PyG \citep{Fey/Lenssen/2019} for the implementation of GNNs and DGL \citep{wang2019dgl} for approximate GED label calculation. Comparison with CVE showcases the abilities of our model-agnostic method compared to theirs, which requires white-box model access and relies on pixel-level edits. On the other hand, comparison with SC demonstrates the power of graph representations compared to set-level edits in the black-box conceptual setting. An important clarification is that SC proposes the use of roles only in the corresponding experiments of \S\ref{sec:extend}, meaning that for \S\ref{sec:cub}, \ref{sec:conceptual} they solely rely on concepts. 
More details in Appendices \ref{sec:setting-details}, \ref{sec:quant}, \ref{sec:qual}.

\subsection{Counterfactuals on CUB}
\label{sec:cub}

\input{sections/cub_experiments}

\subsection{Towards conceptual counterfactuals}
\label{sec:conceptual}
\input{sections/conceptual_experiments}

\subsection{Extendability of graph-based counterfactuals}
\label{sec:extend}
\input{sections/extendability}

\subsection{Efficiency of graph-based counterfactuals}
\label{sec:efficiency}
\input{sections/efficiency_experiments}

%% file: sections/cub_experiments.tex
We experiment with Caltech-UCSD Birds (CUB) \citep{wah2011caltech}, despite its lack of ground truth scene graphs. Nevertheless, they can easily be constructed by leveraging given structured annotations: we create a central node to represent the bird and establish 'has' edges connecting it to its parts. Each part is linked to its respective attributes using edges labeled with the corresponding feature type (color, shape, etc.).
To be consistent with CVE, we use ResNet50 \citep{resnet} as the classifier under explanation.

\paragraph{Quantitative results} 
We examine the agreement between the counterfactuals $I'_{(B)}$ retrieved by each method (CVE, SC, ours) and
% post-hoc explanations 
the ground truth GED.
% , which serves as the golden standard. 
Our approach outperforms CVE on every ranking metric (Tab. \ref{tab:cub-rank-results}). 
As for SC, metrics are only valid for $k=1$ since it produces a single CE instead of a rank. Therefore, P@1 for SC is $0.02$, much lower than ours. 
In addition, we observe that our approach leads to the \textit{lowest number of overall edits}: In Tab. \ref{tab:n_edits-cub-all}, we can see that our method produces about 1 and 2 fewer edits on average for SC and CVE respectively, strengthening the claim that our CEs correspond to the \textbf{minimum number of edits}. 
\input{tables/cub_rank}
\input{tables/cub_edits}

Additionally, our CEs are tied to \textbf{minimum-cost edits}; more specifically, the resulting GED between the query and the retrieved counterfactual scene graph obtain lower GED scores in comparison to both CVE and SC. The related analysis is presented in App.  \ref{sec:avg-ged}.

\paragraph{Qualitative results} 
%\begin{wrapfigure}{l}{7cm}
% \centering
%  \vskip -0.17in 
%  \includegraphics[width=0.8\columnwidth]{figures/poulia_EXTRA.png}
  % \vskip -0.1in
%  \caption{Results for Rusty $\rightarrow$ Brewer Blackbird. \textcolor{magenta}{Color} denotes best results (lower values).}
%  \label{fig:res1}
%  \vskip -0.05in
%\end{wrapfigure}
\begin{figure}[h!]
    \centering
\vskip -0.05in    \includegraphics[width=1\columnwidth]{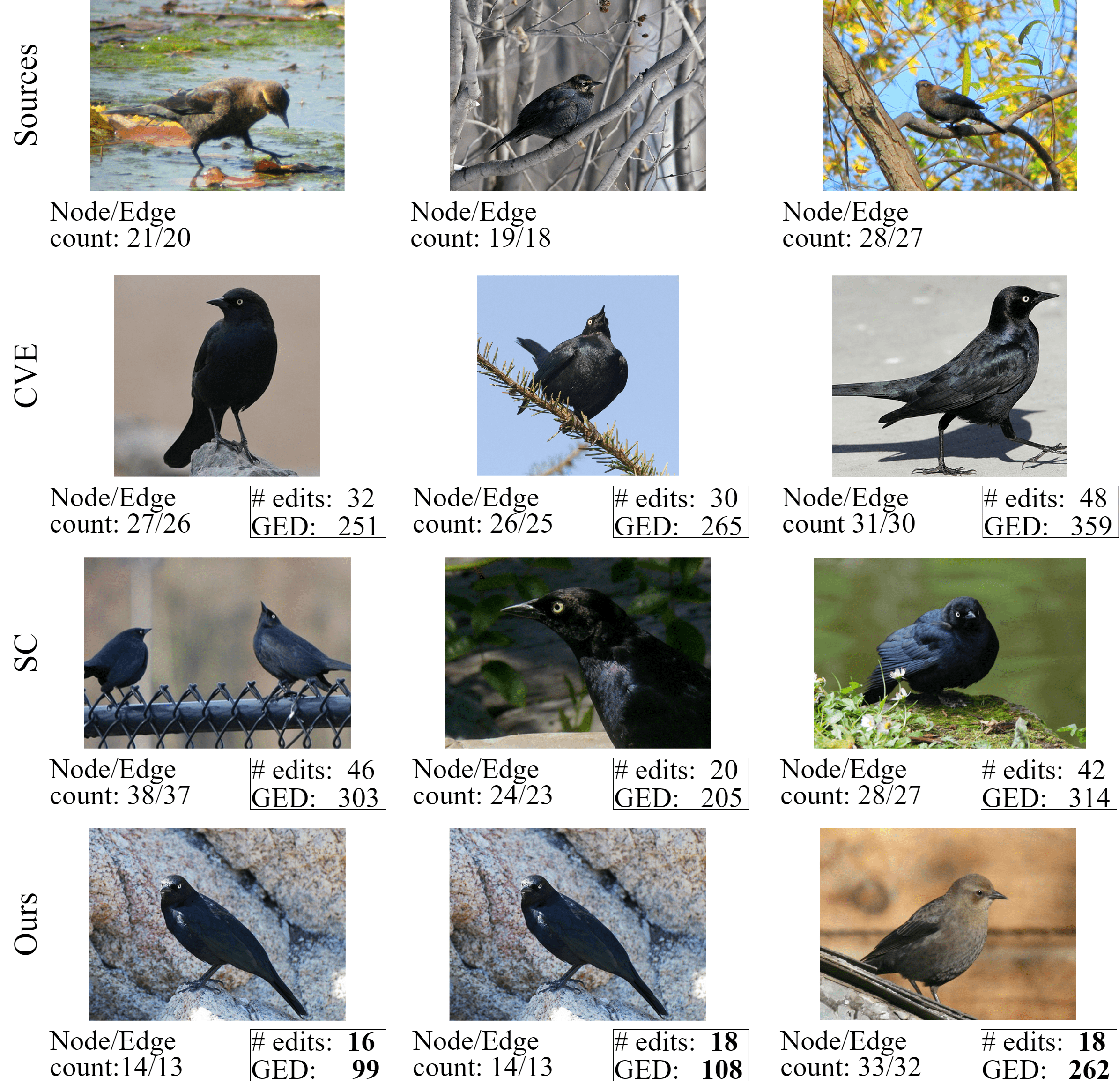}
    \caption{Results for Rusty $\rightarrow$ Brewer Blackbird. \textbf{Bold} denotes best results (lowest number of edits and GED scores).}
    \label{fig:res1}
\end{figure}
for CUB are presented in Fig. \ref{fig:res1} for three images of class A (Rusty Blackbird), accompanied by the number of edits and GED needed to transition to class B (Brewer Blackbird). Overall, our approach produces the \textbf{fewest concept edits}. 
SC leads to clear fallacies like suggesting CEs with additional birds (SC, left), or with a portion of the bird in view (SC, middle); thus leading to unnecessary costly deletions and additions. In contrast, our approach mitigates such errors via graphs, where concept instances are uniquely tied to nodes, and their interconnections strongly guide graph similarity through GED, ultimately producing a more accurate and expressive notion of distance than flat unstructured sets. 
CVE generally fails in finding CEs conceptually similar to the query $I_{(A)}$, as highlighted by the elevated GED and number of edits. Their approach avoids SC's mistakes to an extent by implicitly taking visual features like zoom into account. However, it offers no semantic guarantees, unlike our GED-based approach.

\paragraph{Human evaluation}  
Analyzing the results from the comparative human survey (Tab. \ref{tab:human_eval}), we deduce that our CEs are \textbf{more human-interpretable} than both SC and CVE by a landslide: annotators prefer ours at nearly twice the rate of the CVE alternative.  Compared to SC, despite the increased amount of undecided annotators, our CEs were preferred 2.6 times more frequently. 
This proves that despite the closeness of the two conceptual methods, ours is more intuitive to humans, confirming the meaningful addition of linking concepts within a graph.
A chi-square test revealed significant differences in user preferences between our method and SC (p = 0.003) as well as our method and CVE (p = 9.21e-08), indicating a notable deviation from the expected distribution and further validating the reported results.

As for the machine teaching experiment, we obtain the test set accuracy scores (Tab. \ref{tab:machine-teaching}), as the ratio of correctly human-classified test images over the total number of test images. Our visually-informed accuracy clearly outperforms reported CVE scores, highlighting that concept-based CEs are more powerful in guiding humans towards understanding discriminative concepts between classes compared to non-conceptual pixel-level CEs. The "blind" results show an expected decrease compared to the visually-informed ones, but still outperform CVE. The higher accuracy of concept-based over visual CEs affirms the significance humans place on higher-level features for classification. Details regarding human evaluation are presented in App.  \ref{sec:human}.
\input{tables/cub_human}

\paragraph{Actionability concerns} CVE may lead to non-actionable CEs, despite training on visual semantic preservation. To elaborate, we observe the following: CVE suggests that only adding a striped pattern in a Gray Catbird's wing is adequate to classify it as a Mockingbird. However, by exhaustively generating all annotated attribute combinations of this new bird instance, we easily find several occurring attribute pairs that are not representative of the Mockingbird class; namely, no other Mockingbird has an eyering head pattern and grey breast color.
% This strongly contrasts with CEs retrieved by our method. 
Actionability dictates the prescription of attainable goals achieved through CEs that accurately represent the underlying data distribution \citep{poyiadzi2020face}.
To this end, our approach not only selects CEs drawn from the existing target class distribution but also considers all edits needed to convert query to counterfactual image. Therefore, through GED we formalize a more holistic approach to distance and path between counterfactual pairs and simultaneously leverage relations between depicted objects, both visual (relations on the image) and semantic (relations mapped to WordNet synsets). Further analysis in App. \ref{sec:edits}.
% More details are presented in App.  \ref{sec:edits}.

\paragraph{Global counterfactuals} in terms of \textit{graph edits} require a standardized unit to be changed, in our case referring either to graph triples in a \textit{(concept-edge-concept)} format or merely to \textit{concept edits} as parts of graph triples. Both approaches regard the aggregation of local edits to explain the given classifier from a higher-level perspective. In the case of CUB, global CEs highly correlate with human perception: by considering the  Parakeet Auklet $\rightarrow$ Least Auklet class transition, some key characteristics of the source class (such as the \textit{('beak', 'shape', 'specialized')} triplet) need to be deleted, while others (such as the \textit{('beak', 'shape', 'cone')} triplet) should be added. Further details in App.  \ref{sec:global-cub}.

%% file: tables/cub_rank.tex
\begin{table}[h!]
    \centering
    \caption{Comparison of counterfactual retrieval results with ground truth GED rankings on CUB. \textbf{Bold} denotes best results.}
    \label{tab:cub-rank-results}
    \vskip 0.15in
    \begin{tabular}{P{0.5cm}|P{0.7cm}P{0.7cm}|P{0.7cm}P{0.6cm}|cc}
    \toprule
    & \multicolumn{2}{c}{\small P@k$\uparrow$} & \multicolumn{2}{c}{\small P@k (binary)$\uparrow$} & \multicolumn{2}{c}{\small NCDG@k (bin.)$\uparrow$} \\
    \cline{2-7}
    \small & \small k=1  & \small k=4 & \small k=1  & \small k=4 & \small k=1  & \small k=4 \\
    \hline
    \small CVE & \small 0.02  & \small 0.10 & \small 0.02  & \small 0.11 & \small 0.11  & \small 0.26 \\
    
    \small Ours & \small \textbf{0.19}  & \small \textbf{0.34} & \small \textbf{0.19} & \small \textbf{0.49} & \small \textbf{0.23} & \small \textbf{0.36} \\
    \bottomrule
    \end{tabular}
    \vskip -0.05in
\end{table}

%% file: tables/cub_edits.tex
% \begin{table*}[]
\begin{table}[h!]
\centering
\vskip -0.12in
\caption{Average number of node, edge \& total edits on CUB. \textbf{Bold} for best results (lowest number of edits).}
\label{tab:n_edits-cub-all}
\vskip 0.15in
\begin{tabular}{c|ccc}
\toprule
& \small Node $\downarrow$ & \small Edge $\downarrow$ & \small Total $\downarrow$ \\
\hline
%\hskip -0.08in
\small CVE &  \small 8.43 &  \small 4.70 &  \small 13.13 \\
%\hskip -0.05in
\small SC  &  \small 8.07 &  \small \textbf{3.66} & \small 11.73 \\
%\hskip -0.08in
\small Ours &  \small \textbf{6.16} &  \small 4.34 &  \small \textbf{10.5} \\
\bottomrule
\end{tabular} 
\vskip -0.1in
% \end{table*}
\end{table}

%% file: tables/cub_human.tex
% \begin{table}[h!]
\begin{table}[h!]
% \vskip 2in
\vskip -0.05in
\centering
\caption{Human preference; Win\%=\% times our method was preferred, Lose\% for vice-versa, Tie\% when equally preferred. \textbf{Bold} denotes higher human preference per method.}
\vskip 0.15in
\label{tab:human_eval}
\begin{tabular}{c|ccc}
\toprule
\small Ours & \small Win\% & \small Lose\% & \small Tie\% \\
\hline
\small SC & \small \textbf{48.86} &  \small 19.32 &  \small 31.82 \\
\small CVE & \small \textbf{48.42} &  \small 26.27 &  \small 25.31\\
\bottomrule
\end{tabular}
\end{table}

\begin{table}[h!]

% \begin{wraptable}{r}{5.5cm}
\centering
\vskip -0.08in
\caption{Human test accuracy scores for correct classification of samples in classes A and B. \textbf{Bold} for top accuracy score.}
\label{tab:machine-teaching}
\vskip 0.15in
\begin{tabular}{c|c}
\toprule
\hspace{-5px}\small Human experiment & \small Test accuracy \%$\uparrow$ \\
\hline
\hspace{-5px}\small Ours (visually-informed)  & \small \textbf{93.88}\\
\small Ours (blind) & \small 89.28 \\
\small CVE & \small 82.1 \\
\bottomrule
\end{tabular}
\vskip -0.1in
\end{table}

%% file: sections/conceptual_experiments.tex
We focus our analysis on conceptual counterfactuals since the previous sections exhibited the indisputable merits of such approaches against the SotA pixel-level method of CVE. In the interest of experimenting on a less controlled dataset, we employ Visual Genome (VG) \citep{krishna2017visual}, a dataset containing over 108k human-annotated scene graphs, describing scenes of multiple objects and their in-between interactions. We construct two manageable subsets of 500 scene graphs each, corresponding to $\sim$125k possible training graph pairs for our GNNs. The first subset denoted as VG-RANDOM is randomly selected, while the second one, named VG-DENSE, is chosen to favor higher graph densities and less isolated nodes to highlight the importance of object interconnections. Details are provided in App. \ref{sec:graph_stats}.
VG instances lack ground truth classification labels, allowing us to test our counterfactual retrieval method without the definition of a certain target class.
We classify instances using a pre-trained Places365 classifier \citep{places}, and regard as counterfactual classes the closest ones in rank. Specifically, we employ a pre-trained ResNet50 \cite{resnet}, as proposed in the original Places365 paper.

\paragraph{Quantitative Results}
We first compare the average number of edits for our method and SC (Tab. \ref{tab:n_edits-vg}). Initially, numerical results between the two methods seem similar, but upon closer inspection in conjunction with average GED results of Tab. \ref{tab:avg-ged-vg}, our method's superiority is evident. VG contains concepts which are much more diverse than CUB, and despite the knowledge-based contraints we enforced during GED computation, edit distance is expected to be higher \textit{between concepts}. Note that this is not true for mean GED since CUB has a higher number of average edits. To this end, our method leads to lower GED in all cases, even when number of edits is higher (VG-RANDOM). Some extra analysis is provided in App.  \ref{sec:avg-ged}.
\input{tables/vg_edits}

\input{tables/vg_avg_ged}

Regarding CE approximation to ground truth GED, results for our approach are presented in Tab. \ref{tab:hit-percentage-vg} denoted as GCN-70K.
As for SC, P@1 is 0.25 on VG-DENSE and 0.20 on VG-RANDOM, compared to 0.25 and 0.21 retrieved by our method. GED approximation is satisfactory but close between methods due to a general agreement in CE retrieval.
% (App.  \ref{}). 
Despite this fact, our approach still leads in all reported metrics, especially for VG-DENSE, displaying superiority in cases of disagreement. 
Ranking results consistent with our analysis are also obtained using GQA \cite{hudson2019gqa}, a VG variant which focuses on question-answering, as reported in App. \ref{sec:quant-additional}. 
Reported findings 
% finding
% The closeness of overall results 
place great significance in examining qualitative results.
\input{tables/vg_rank}
% for this experiment.
% ; therefore, the following qualitative results demonstrate the advantages of our approach. 

\begin{figure*}[t!]
  %\hspace{-15px}
  \centering
\vskip -0.01in
\includegraphics[width=0.9\textwidth]{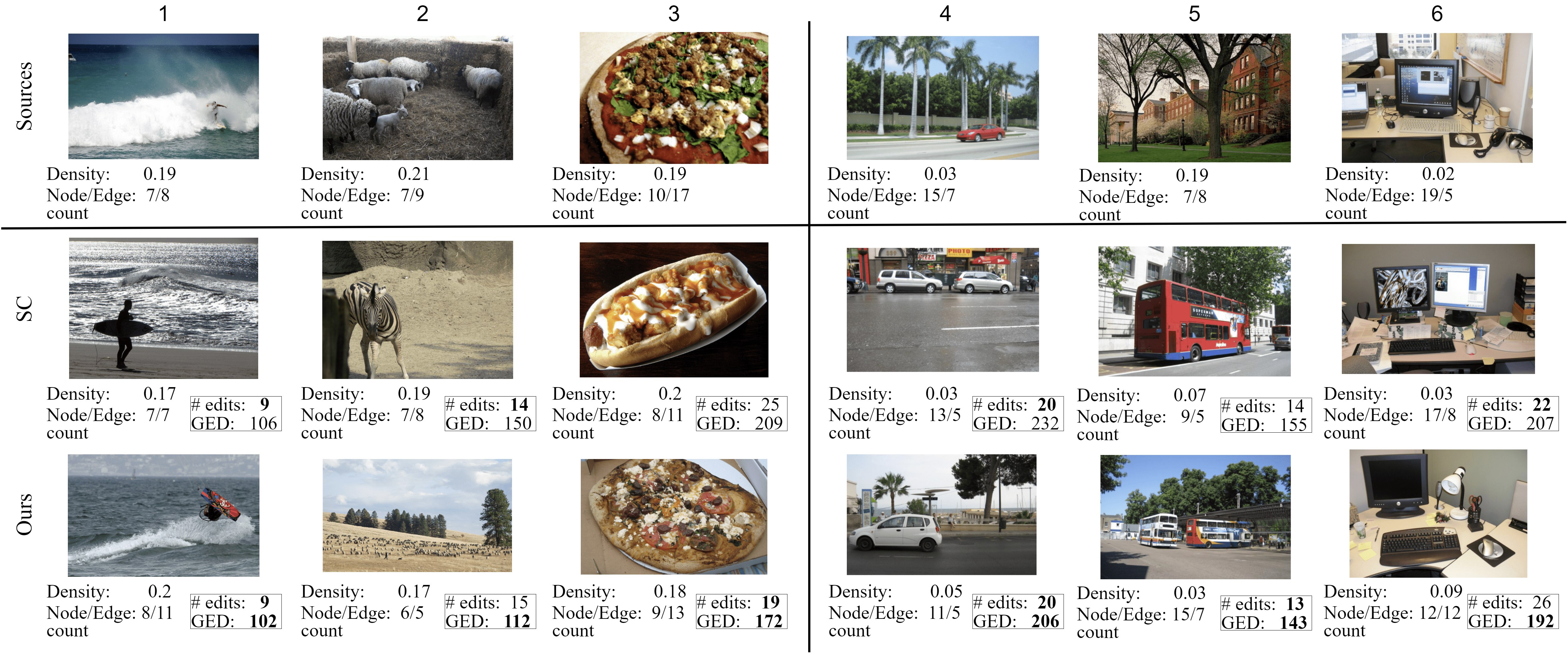}
  \caption{Qualitative results (best metrics in \textbf{bold}): VG-DENSE (left 3 columns) and VG-RANDOM (right 3 columns). }
\label{fig:res2}
\vskip -0.06in
\end{figure*}

\paragraph{Qualitative results}
By examining counterfactual images retrieved for VG-DENSE in Fig. \ref{fig:res2} (left), there is 
a clear indication that by considering the complex relations between concepts, our method achieves more 
\textbf{detail-oriented results}: in the 1st column, our approach not only retrieves an image
with 'man', 'board', 'water' concepts, but also %contains 
the relation 'man on board'. 
In the 3rd column, we consider the relation of toppings and retrieve the pizza, while SC simply retrieves an image with similar concepts, ('bun' and 'bread' or 'meat' and 'sausage').  
Results on VG-RANDOM (Fig. \ref{fig:res2} (right)) follow the same logic. In columns 4-5, our method retrieves the focal
points of the images since it regards relations between trees and other objects. Taking into account the sparsity of the underlying graphs, however, in some cases
the importance of concepts trumps the underlying structure, as in the 6th column. This fact is reflected
in the elevated number of edits of our method for VG-RANDOM, yet it is not true for GED, showcasing once again the importance of semantic context. More details in App.  \ref{sec:graphs_vg}.

\paragraph{Why GCN?} Ranking metrics of GNN models are provided in Tab. \ref{tab:hit-percentage-vg}. Three GNN variants (GAT, GIN, GCN) are trained using $p=N/2=$70k scene graph pairs.
The GCN-based variant consistently approaches GED the closest, with a binary P@4 of 49\% and P@1 of 24.80\% for VG-DENSE and slightly worse results on VG-RANDOM.  GCN systematically scores higher in comparison to theoretically more competent GNN alternatives, such as GIN. We attribute this finding to the importance of local neighborhood information for our small yet semantically dense graphs. Specifically, the VG graphs considered in our experiments rarely exceed 3-4 hops, as briefly demonstrated in App. \ref{sec:graphs_vg}. GIN does not incorporate node features during aggregation resulting in a limited notion of semantic similarity. 
This ablation study affirms using GCN for the GNN-based similarity component of our approach. GNNs can also outperform other prominent deterministic methods, like graph kernels \citep{grauman2007pyramid}.
The reported findings grant us the security that our counterfactual explanations are \textbf{trustworthy}, even when applied to complex scene graphs.

%% file: tables/vg_edits.tex
\begin{table}[h!]
% \begin{wraptable}{r}{8.4cm}
% \vskip -0.17in
\centering
\vskip -0.05in
\caption{Average number of node, edge \& total edits on VG.
\textbf{Bold} denotes best results (lowest number of edits).
}
\label{tab:n_edits-vg}
\vskip 0.15in
\begin{tabular}{P{0.5cm}|ccc|ccP{0.6cm}} 
\toprule
&  \multicolumn{3}{c}{\small VG-DENSE} &  \multicolumn{3}{c}{\small VG-RANDOM} \\
\cline{2-7}
& \small Node$\downarrow$ & \small Edge$\downarrow$ & \small Total$\downarrow$& \small Node$\downarrow$ & \small Edge$\downarrow$ & \small Total$\downarrow$\\
\hline
\small SC  &  \small \textbf{4.91} &  \small 7.29 &  \small 12.2 &  \small \textbf{12.15} &  \small \textbf{7.52} &  \small \textbf{19.67} \\
\small Ours &  \small 4.95 &  \small \textbf{7.15} &  \small \textbf{12.11} &  \small 12.18 &  \small 7.54 &  \small 19.72\\
\bottomrule
\end{tabular} 
% \vskip -0.12in
% \end{wraptable}
\end{table}

%% file: tables/vg_avg_ged.tex
% \begin{table*}[h!]
\begin{table}[h!]
% \vskip -0.17in
    \centering
    %\vskip -0.23in
        \caption{Average top-1 GED (VG) for CEs when methods disagree. \textbf{Bold} for best (lowest) GED scores for each dataset split.
    }
    \label{tab:avg-ged-vg}
    \vskip 0.14in
    \begin{tabular}{c|c|c} 
        \toprule
        & \small VG-DENSE $\downarrow$ & \small VG-RANDOM $\downarrow$ \\
        \hline
        \small SC   & \small 128.67  & \small 186.77 \\
        \small Ours  & \small \textbf{122.41} & \small \textbf{180.67} \\
        \bottomrule
    \end{tabular} 
    \vskip -0.1in
\end{table}

%% file: tables/vg_rank.tex
\begin{table}[h!]
\vskip -0.05in
\begin{center}
%\hspace{-0.4em}
\caption{Ranking results on the two VG variants for various GNNs. \textbf{Bold} numbers indicate best ranking metrics.}
\label{tab:hit-percentage-vg}
\vskip 0.01in
\begin{tabular}{P{1.35cm}|p{0.65cm}p{0.65cm}|p{0.67cm}p{0.67cm}|p{0.5cm}p{0.4cm}}
\toprule
& \multicolumn{2}{c}{\small P@k $\uparrow$} & \multicolumn{2}{c}{\small P@k (binary)$\uparrow$} & \multicolumn{2}{C{1.6cm}}{\small NDCG@k (binary)$\uparrow$}\\  
\cline{2-7}
\small Models & \small k=1  & \small k=4 & \small k=1  & \small k=4 & \small k=1  & \small k=4 \\
\hline
& \multicolumn{6}{c}{\small VG-DENSE}\\
 \cline{2-7} \small Kernel& \small 0.13& \small 0.17& \small 0.13& \small 0.26& \small 0.19&\small 0.33\\
\small GIN-70K& \small 0.16  &  \small 0.27 &  \small 0.16  &  \small 0.38 &  \small 0.20  &  \small 0.34 \\
\small GAT-70K & \small 0.18  &  \small 0.32 &  \small 0.18  &  \small 0.44 &  \small 0.22  &  \small 0.35 \\
\small GCN-70K & \small \textbf{0.25}  &  \small \textbf{0.37} &  \small \textbf{0.25}  &  \small \textbf{0.49}  &  \small \textbf{0.28} &  \small \textbf{0.41}  \\
\hline
& \multicolumn{6}{c}{\small VG-RANDOM}\\
 \cline{2-7} \small Kernel& \small 0& \small 0.01& \small 0& \small 0.01& \small 0.10&\small 0.25\\
\small GIN-70K& \small 0.03  &  \small 0.07 &  \small 0.03  &  \small 0.07 &  \small 0.22  &  \small 0.38 \\
\small GAT-70K & \small 0.18  &  \small 0.29 &  \small 0.18  &  \small 0.38 &  \small 0.11 &  \small 0.27 \\
\small GCN-70K & \small \textbf{0.21}  &  \small \textbf{0.30} &  \small \textbf{0.21}  &  \small \textbf{0.42}  &  \small \textbf{0.25}  &  \small \textbf{0.38} \\
\bottomrule
\end{tabular}
\end{center}
\vskip -0.1in
\end{table}

%% file: sections/extendability.tex
The flexibility of our approach is proven under two 
%separate 
scenarios: a) its application on unannotated images,
b) its expansion into other modalities. For direct comparison to SC we provide global CEs by averaging overall graph triple edits.
% (additions, deletions, substitutions).

\paragraph{Unannotated datasets} 
% ONLY EXPERIMENT SC WITH EDGES / roles

\begin{figure}
  \centering
  % \vskip -0.2in 
  % \hskip -0.1in
\includegraphics[width=0.48\textwidth]{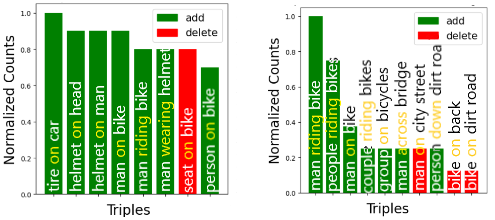}
  \caption{Graph edits (triples inserted/ deleted) to implement the 'pedestrian' $\rightarrow$ 'driver' transition. The yellow color distinguishes edge from node labels within a triple.}
  \label{fig:sgg}
   % \vskip -0.08in
\end{figure}

We replicate \citet{wisely}'s experiment
on explaining the classification of web-crawled creative-commons images into 'driver' and 'pedestrian' classes. Here, images were manually classified by the authors; thus, we explain a non-neural classifier.
% (App.  \ref{sec:limitations}). 
By employing the SotA scene graph generator (SGG) of \citet{reltr} we extract global edits from generated graphs for the transition from 'pedestrian' to 'driver' (Fig. \ref{fig:sgg}(left)). 
Their relevance is verified by our common sense: people wear helmets when driving -addition of (helmet, on, head) and (man, on, bike)- and cover the bike seat with their body -deletion of (seat, on, bike)-.
To validate our method's consistency across other annotation techniques, we replace the SGG with a pipeline of captioning (BLIP \citep{blip}) and graph parsing (Unified VSE \citep{Wu2019UnifiedVE}).
We confirm that resulting edits (Fig. \ref{fig:sgg} (right)) semantically resemble the ones in Fig. \ref{fig:sgg} (left). Overall, similarly to Fig. \ref{fig:example1}, more accurate local edits are achieved through the consideration of the multiplicity of objects and relations. Generic triple edits result from errors in the automatic annotation pipeline, emphasizing the importance of meticulous explanation dataset curation.
We provide further details regarding the 'pedestrian' vs 'driver' classification experiment in App.  \ref{sec:unannotated}.
Additionally, we experiment with more unannotated datasets of scene images, such as Action Genome \cite{ji2020action}, which is presented in App.  \ref{sec:unannotated}.

\paragraph{Audio classification} 
Despite focusing on images, we briefly demonstrate our method's model-agnostic nature by applying it for audio classification, following SC. 
We provide CEs using the Smarty4covid dataset \citep{smarty} for the IEEE COVID-19 sensor informatics competition winner\footnote{\href{https://healthcaresummit.ieee.org/data-hackathon/ieee-covid-19-sensor-informatics-challenge/}{IEEE COVID-19 sensor informatics competition}}, which predicts COVID-19 from cough audio. 
Our findings align with SC, revealing the high frequency of concept edits among respiratory symptoms and uncovering the same gender bias.
No new findings are produced for this primarily concept-based dataset with trivial interconnections, once again placing the focus on the nature and density of the annotations. However, we confirm that our method is at least as good as SC in these cases nonetheless. More results regarding the audio classification experiment are discussed in App.  \ref{sec:smarty}.

% particularly respiratory symptoms in Tab. \ref{table:covid}, and uncovers the reported gender bias of the training dataset which includes more COVID-positive women than men. Note that presented edits are all additions and in the form ('User', 'symptom', X) where X is an element of Tab. \ref{table:covid}. A longer list is in App.  \ref{sec:smarty}.

%% file: sections/efficiency_experiments.tex
\paragraph{Time Performance for Counterfactual Retrieval}
From a theoretical standpoint, our method's efficiency is expected. The heaviest part of GNN computations occurs during model training, where each backward call is correlated with the square of the number of nodes in addition to the number of edges. Inference, on the other hand, is nearly instantaneous and is linearly correlated with the sum of the number of nodes and edges in a graph \cite{gnn-complexity}.

Additionally, we experimentally confirm that our method allows for \textbf{efficient} CE retrieval: In Tab. \ref{tab:time}, we report execution times for CE computation on the complete sets of graphs 
%from our experiments, 
using GED 
\citep{fankhauser2011speeding} 
versus our GNN-powered approach. We further report retrieval and inference time of our method. 
Even by adding times for all GCN-N/2 operations, we significantly relieve the computational burden of calculating the ground truth GED for all graph pairs, especially for larger graphs.
\input{tables/perf}

\begin{figure}[htb]
  \centering
  \begin{subfigure}[b]{1\columnwidth}
    \centering
    \includegraphics[width=1\columnwidth]{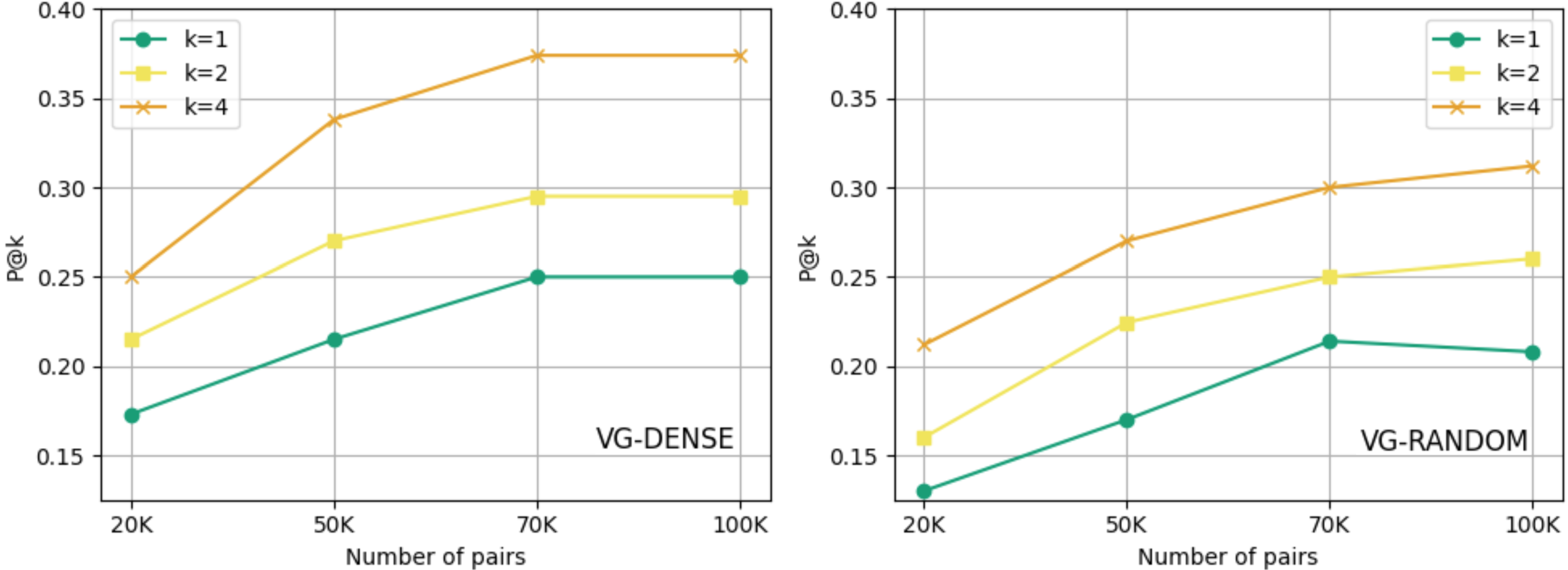}
    \caption{VG variants}
    \label{fig:comparison-dense}
  \end{subfigure}

  \begin{subfigure}[b]{1\columnwidth}
    \centering
    \vskip 0.1in
    \includegraphics[width=0.5\columnwidth]{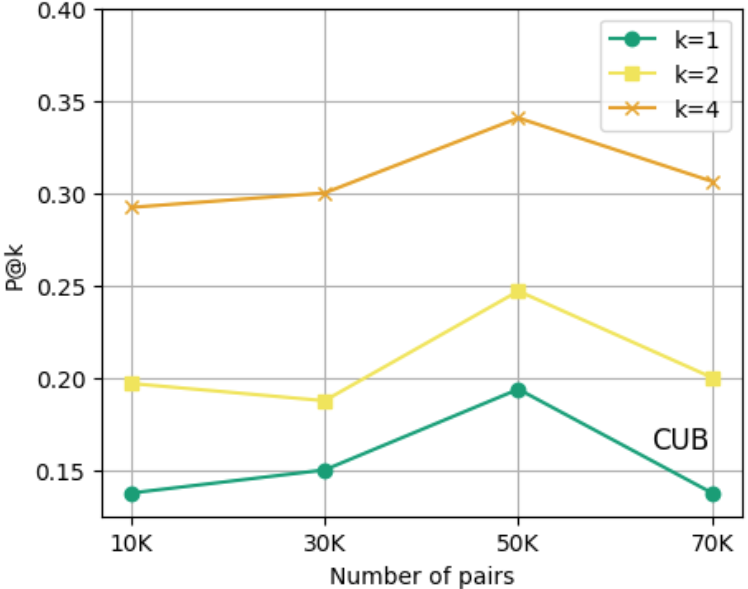}
    \caption{CUB}
    \label{fig:comparison-cub}
  \end{subfigure}

  \caption{P@k of GCN variant for different training pairs $p$ on the two main datasets explored.}
  \label{fig:comparison}
\end{figure}

\paragraph{Performance-complexity trade-off}
In Fig. \ref{fig:comparison}, we examine how retrieval precision varies using different numbers of training pairs $p$ on CUB (Fig. \ref{fig:comparison-cub}) and the two VG variants (Fig. \ref{fig:comparison-dense}). P@k does not exhibit significant increase in any case after the $\sim N/2$ pairs mark (70k for VG and 50k for CUB). On the contrary, it could remain identical (Fig. \ref{fig:comparison-dense} left) or even decrease (Fig. \ref{fig:comparison-cub}). The same precision pattern per $p$ is experimentally validated on the GQA dataset (Appendix \ref{sec:quant-additional}). The consistency of behavior exhibited over $\sim N/2$ pairs concludes our claim that \textbf{$N/2$ training pairs are adequate} for appropriate graph embedding using GCN.
% \begin{figure}[htb]
%   \begin{minipage}{0.49\columnwidth}
%     \hspace{-10px}
%     \includegraphics[width=1.06\textwidth, height=0.8\textwidth]{figures/vg-dense-pairs.png}
    
%   \end{minipage}
%   \begin{minipage}{0.5\columnwidth}
%   \hspace{18px}
%     \includegraphics[width=1.07\textwidth, height=0.78\textwidth]{figures/vg-random-pairs.png}
%     % \caption{VG-RANDOM}
%     % \label{fig:comparison-random}
%   \end{minipage}
%   \caption{P@k of GCN variant for different training pairs $p$.}
%   \label{fig:comparison-dense}
% \end{figure}

%% file: tables/perf.tex
\begin{table}[h!]
\vskip -0.15in
\caption{Time (sec) for counterfactual calculation. Training time is reported due to the transductivity of the GNN method. }
\label{tab:time}
\vskip 0.15in
\begin{tabular}{C{2cm}|C{0.4cm}C{1.25cm}|C{1.25cm}C{1.25cm}}
    \toprule
    & \small GED \hfill \hfill \hfill $\downarrow$ & \small GCN-N/2 (train)$\downarrow$ & \small GCN-N/2 (retr.)$\downarrow$ & \small GCN-N/2 (infer.)$\downarrow$    \\
    \hline
    \small CUB        & \small  46220 & \small 32691 & \small 0.03 & \small 0.06   \\
    \small VG-DENSE   & \small 13982  & \small 12059 & \small 0.03 & \small 0.06   \\
    \small VG-RANDOM  & \small 18787  & \small 16271 & \small 0.03 & \small 0.10   \\
    \bottomrule
\end{tabular}   
\end{table}

%% file: sections/conclusion.tex
\section{Conclusion}
\label{sec:conclusion}
% could have discussion
In this paper, we proposed a new model-agnostic approach for counterfactual computation based on the expressive power of semantic graphs. To this end, we suggested counterfactual retrieval by GED calculation, employing 
a GNN-based similarity model to accelerate the otherwise NP-hard retrieval process between all input graph pairs.  
Comparison with previous CE models proved that
our explanations correspond to minimal edits and are more human interpretable, especially when interactions between concepts are dense, while still ensuring actionability. 
We further confirmed the applicability of our framework on datasets without annotations. 
There is ample room for future work, including exploring potential limitations,
% of CE methods, 
like robustness and the impact of low quality annotations, as well as further improving the efficiency 
by employing unsupervised GNN methods.

%% file: sections/impact.tex
\section*{Acknowledgments}

This work has been developed as part of the HiDALGO2 project, which has received funding from the European High Performance Computing Joint Undertaking (JU) and Poland, Germany, Spain, Hungary, France and Greece under grant agreement No 101093457.
Maria Lymperaiou was supported by the Hellenic Foundation
for Research and Innovation (HFRI) under the 3rd Call for HFRI PhD Fellowships
(Fellowship Number 5537). 
We thank all reviewers for their insightful comments and feedback, and our annotators for participating in the conducted human surveys.

\section*{Impact Statement}
This paper presents work whose goal is to advance the field of Explainability in Machine Learning. There are no societal consequences of our work concerning the utilization of data such as image datasets or annotations. However, end users should exercise caution when relying on explanation methods, not specifically ours but generally, especially in sensitive domains. This caution stems from the possible presence of low-quality data in the real-world environment, a factor beyond the focus of our research. Users of this system must carefully select their data, as is prudent in any AI application.

%% file: sections/appendix.tex
\section{Human evaluation details}
\label{sec:human}

\subsection{Participants and Consent}
We distributed an information sheet describing the goals and stages of our human surveys to software engineering students online. We clarified that their participation would be voluntary and without any form of compensation. We additionally distributed the following form to obtain annotators' consent in the form of a checklist. We used the same form both for the machine teaching as well as the counterfactual preference experiment. The 33 people who ultimately participated were young adults of ages 19-25 both male and female, without any knowledge of bird species. 

\begin{figure}[h!]
    \centering
    \includegraphics[width=0.65\textwidth]{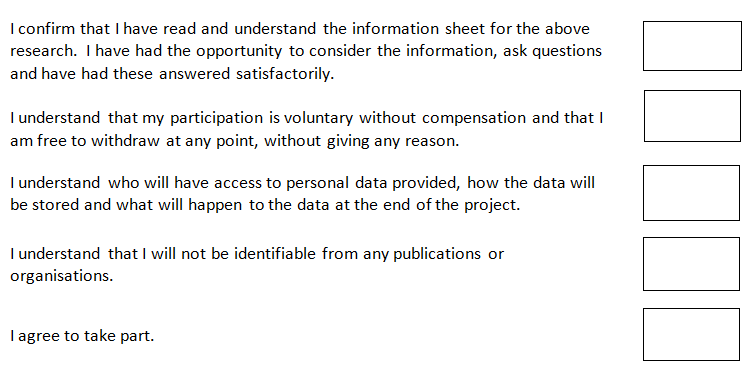}
    \caption{Screenshot of the consent form for human evaluation. Our annotators fill out this form before they proceed with annotations.}
    \label{fig:consent}
\end{figure}

Our human survey was completely anonymous and we did not record any type of personal data from our annotators.

\subsection{1st experiment: comparative human survey}
\begin{figure}[t!]
    \centering
    \includegraphics[width=0.8\textwidth]{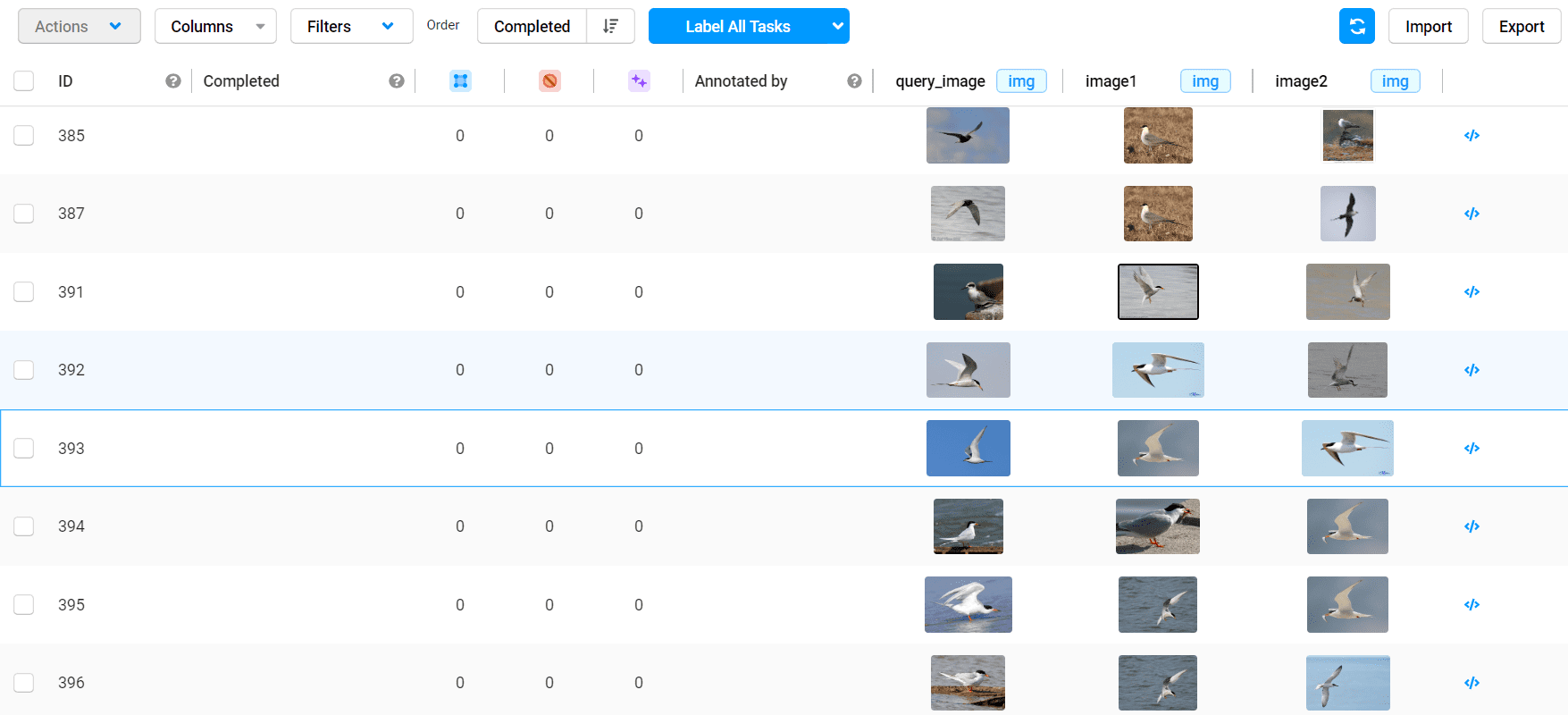}
    \caption{Screenshot of the platform provided for human evaluation.}
    \label{fig:platform}
\end{figure}

\begin{figure}[h!]
    \centering
    \includegraphics[width=0.8\textwidth]{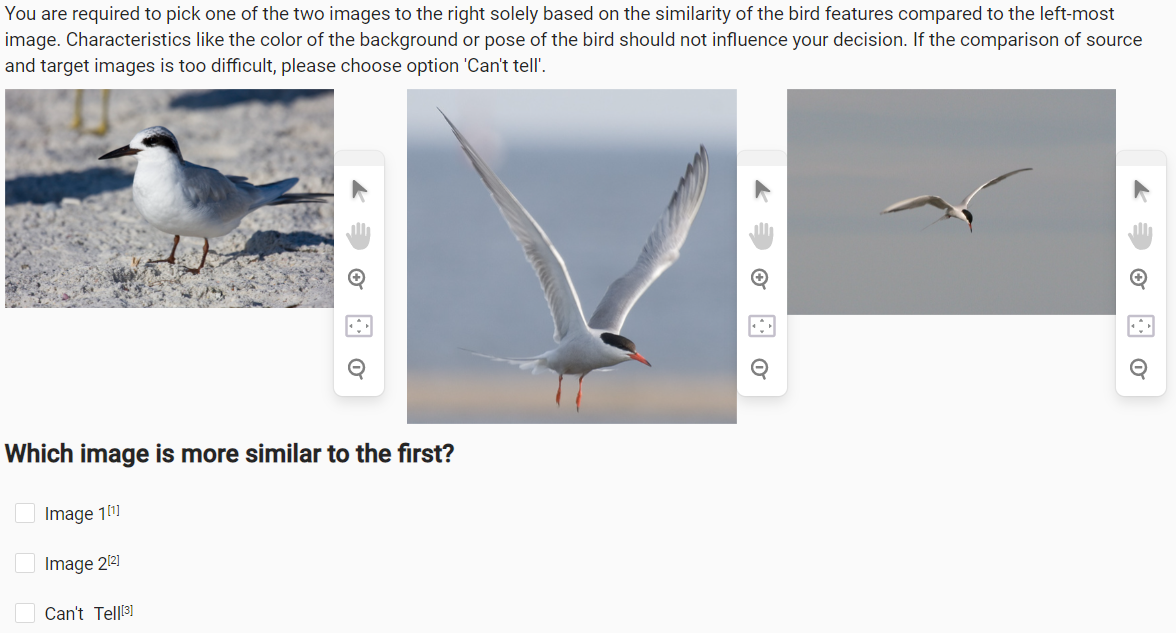}
    \caption{Annotation panel with instructions and image navigation tools provided to the evaluators for CUB.}
    \label{fig:cub-human}
\end{figure}

In Fig. \ref{fig:platform}, we present a screenshot of the platform we provided to our evaluators for the comparative user survey. Users are asked to select a sample to annotate, as shown in the panel of Fig. \ref{fig:cub-human}. We ensured that our evaluators can clearly view the images and their details by providing 'zoom-in'/'zoom-out' tools, as well as the ability to navigate within the image with the 'pan' and 'move' options. 
\newline

An annotator can click on any sample to be annotated, thus moving to a screen such as the one of Figure \ref{fig:cub-human}. The source image is presented on the left, and the two alternative options (ours versus a counterfactual image of CVE \citep{vandenhende2022making} or SC \citep{wisely}) are placed in the middle and the rightmost column. These options are shuffled in each sample, so that no bias towards each choice is created.
% , in cases an annotator chooses to evaluate more than one sample. 
Only one of the options ("Image 1", "Image 2" or "Can't tell") can be selected for each sample. 

In this first human experiment, our annotators can evaluate as many samples as they wish; however, they cannot update an existing annotation. All 33 annotators participated in this experiment.

\subsection{2nd experiment: machine-teaching human survey} 
We once again employ the same platform as for the previous human experiment. However, this time each annotator can only evaluate \textbf{one} single sample; we enforce this restriction to clearly evaluate the contribution of the learning phase, excluding situations that an annotator could have become more 'competent' after passing many times through the learning phase. 

The experimental workflow is adopted from \cite{vandenhende2022making}, therefore we include all the three stages (pre-learning, learning and testing).

\begin{figure}[h!]
    \centering
    \includegraphics[width=0.8\textwidth]{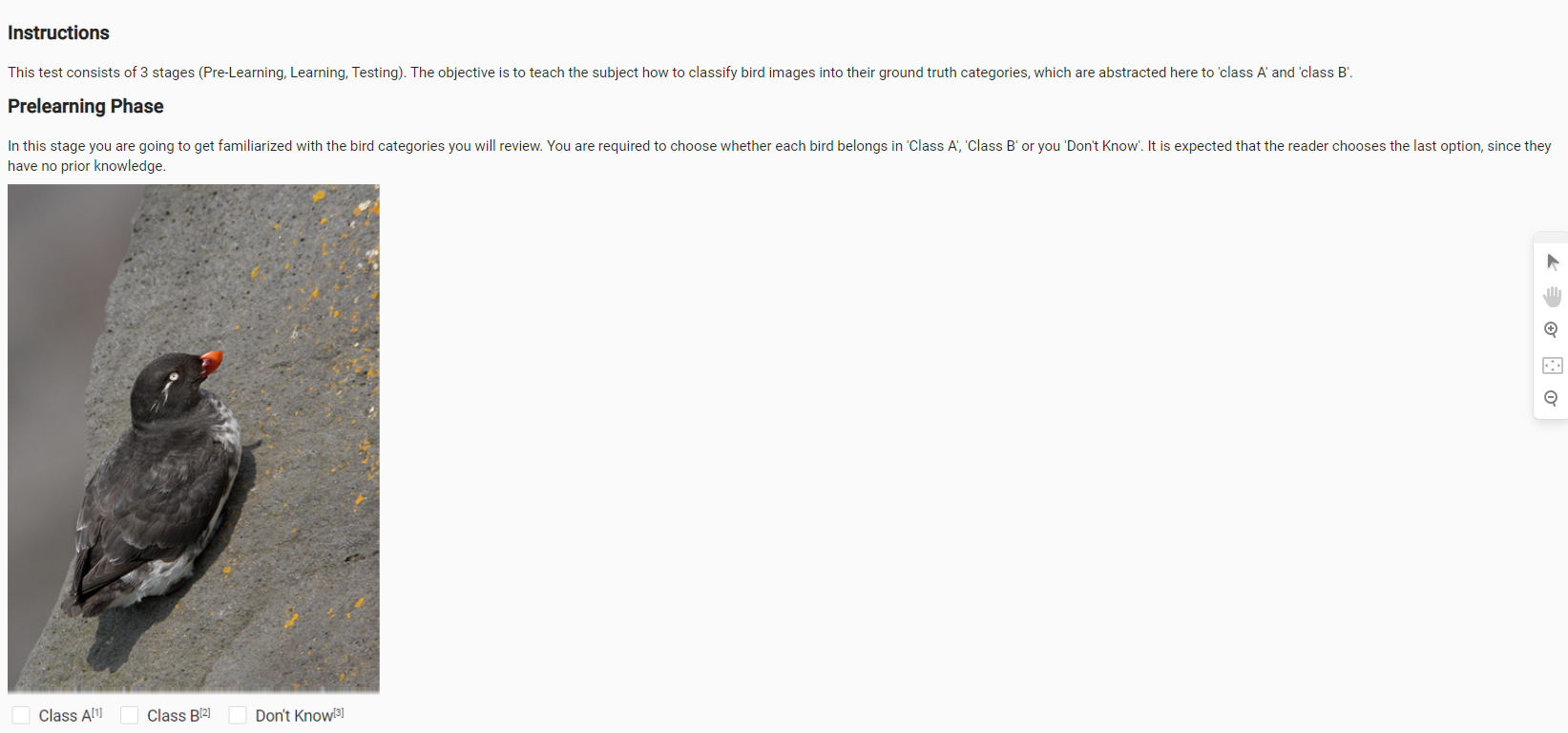}
    \caption{Pre-Learning stage instructions for CUB machine teaching experiment. Choices are "Class A", "Class B" or "I don't know".}
    \label{fig:prelearning}
\end{figure}

\paragraph{Pre-learning stage}
In the pre-learning stage, users are presented with unlabeled images from the test set to get familiarized with the nature of the images they will be tasked to classify later on. Fig. \ref{fig:prelearning} is provided as an example of the pre-learning screen. The annotators become aware that the classification to the anonymized classes A and B cannot be performed without passing through the learning stage, therefore selecting "I don't know" is the expected option.
In Fig. \ref{fig:prelearning}, we can explicitly see the three options for image classification, namely "Class A", "Class B" or "I don't know". Only one can be selected at a time, as in \cite{vandenhende2022making}. 

\paragraph{Learning stage}
The learning stage comprises the heart of this human experiment. As mentioned in the main paper, we perform two variants of it to measure the degree of reliance on concepts, according to human perception. A user can either participate in the "visually-informed" or the "blind" experiment, but not both. This is necessary so that we exclude the possibility of evaluating the same data sample in each of the experiments and thus eliminate the possibility of having some knowledge transfer across the two variants of this experiment. Annotators are divided into equal subgroups (17 in the "visually-informed" variant and 16 in the "blind" one).

\begin{figure}[]
\vspace{-2px}
    \centering
    \includegraphics[width=0.68\textwidth]{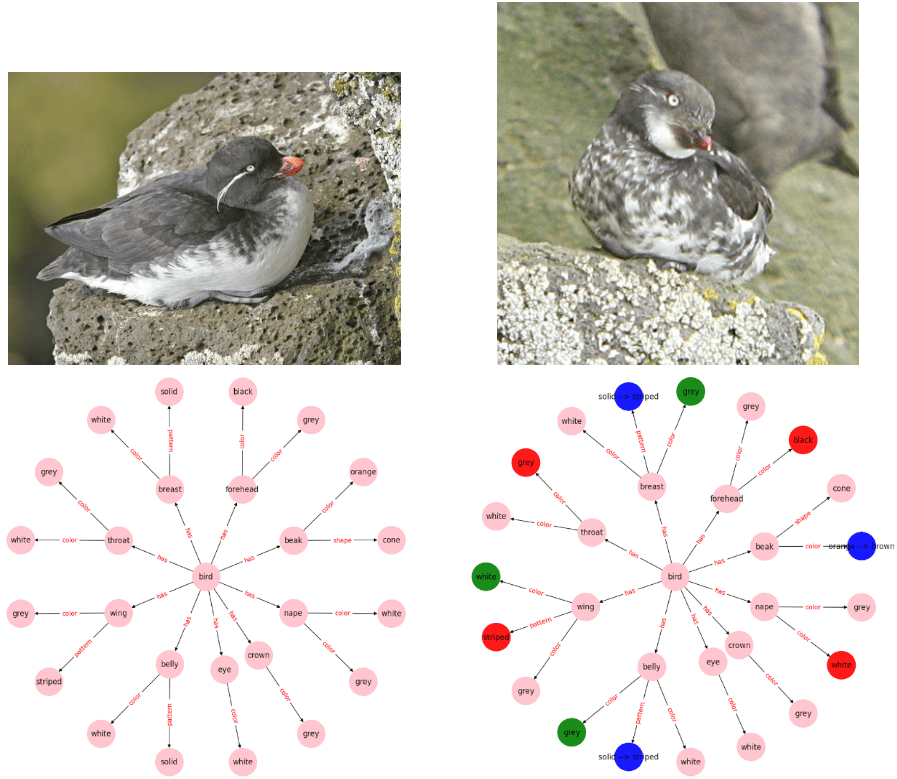}
    \caption{Example of the visually-informed learning stage.}
    \label{fig:learning1}
\end{figure}

\begin{figure}[]
\vspace{-2px}
    \centering
    \includegraphics[width=0.68\textwidth]{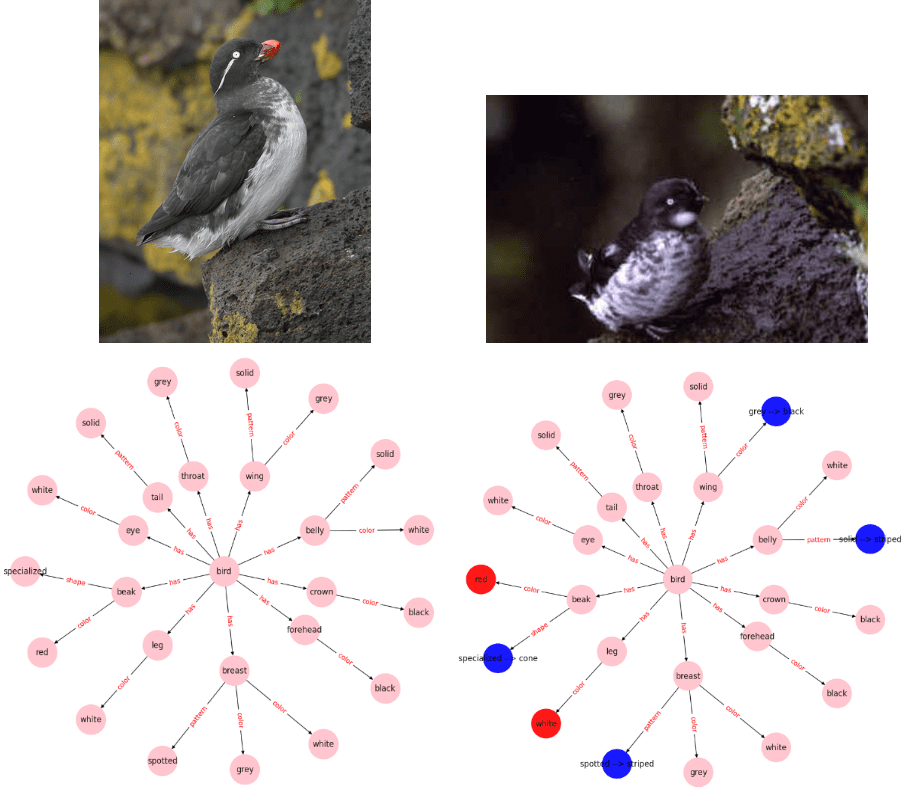}
    \caption{Example of the visually-informed learning stage.}
    \label{fig:learning2}
\end{figure}

\begin{figure}[h!]
\vspace{-2px}
    \centering
    \includegraphics[width=0.7\textwidth]{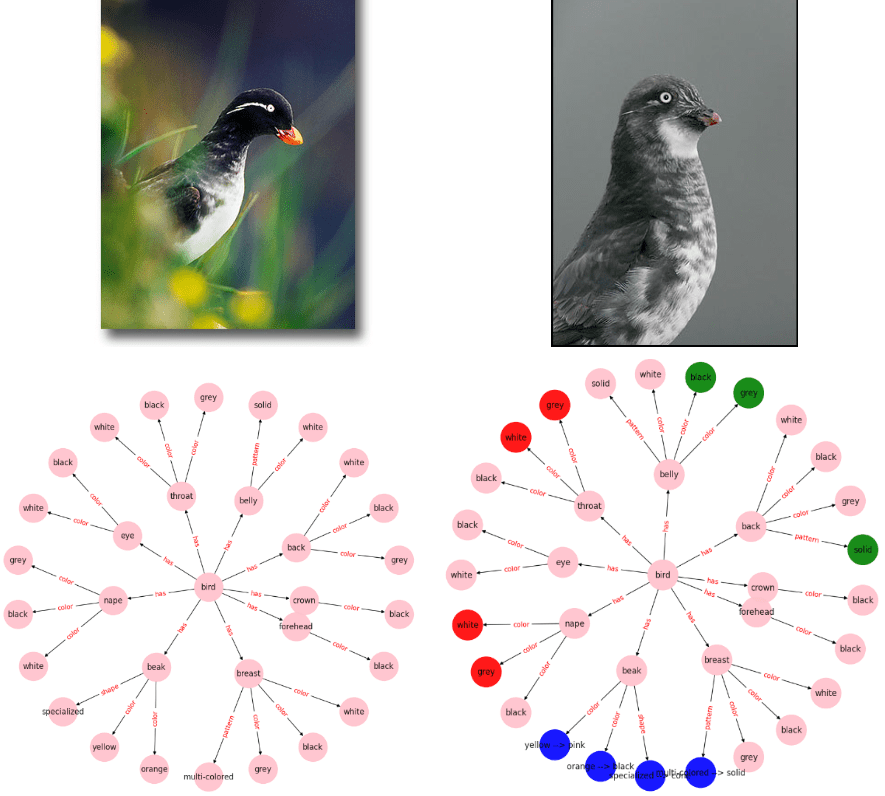}
    \caption{Example of the visually-informed learning stage.}
    \label{fig:learning3}
\end{figure}

In the \textbf{visually-informed} variant, annotators are presented with training images from anonymized classes A and B, together with their scene graphs, as shown in Figures \ref{fig:learning1}, \ref{fig:learning2}, \ref{fig:learning3}. Of course, training and test images do not overlap. Annotators are again provided with 'zoom-in'/'zoom-out', 'pan', 'move' tools, etc. to navigate within the images and the accompanying scene graphs.

Training images on the left always belong to class A, while images on the right always belong to class B. Scene graphs on the right also contain the edits needed to perform the $A\rightarrow B$ transition, with green nodes representing concept additions, blue nodes indicating concept substitutions (both source and target concepts of the substitution are shown), and red nodes denoting concept deletions. The rest of the nodes imply that the corresponding concepts remain the same between the two classes. 

A user implicitly focuses on the most frequent insertions, substitutions, and deletions performed throughout the training stage to understand the discriminative features between class A and class B. Associating such concepts with the images helps mapping graph edits to visual differences so that the user learns to separate classes visually and conceptually.

In the \textbf{blind} variant of the learning stage, only scene graphs are provided, but no training images. Also, the graph edits are presented to the users via colored nodes. This learning variant is a direct analogy to the machine-teaching learning stage implemented by \citet{vandenhende2022making}: in their case, pixels corresponding to discriminative regions that act as explanations are provided, while the rest of the bird image is blurred out. Therefore, annotators need to learn solely from the explanation and mentally connect the corresponding concepts to existing visual regions of the testing images. In our case, the derived explanations correspond to graph edits, therefore annotators have to learn the discriminative concepts that are added, substituted, or deleted to perform the $A \rightarrow B$ transition.
However, since our learning setting is performed without any visual clue, we regard our blind learning stage as being \textbf{more difficult} than the learning stage that \cite{vandenhende2022making} implement; our annotators have to connect concepts with image regions, thus performing cross-modal grounding in order to learn discriminative features.

Throughout the blind learning stage, we are able to measure the reliance on concepts rather than pixels to learn to classify images of unknown classes. This experiment is important in order to highlight how meaningful and informative conceptual explanations are to humans, so that they can approximate a zero-shot classification setting. 

\paragraph{Testing stage} In the testing stage, users are provided with the same images as in the pre-learning stage. No scene graphs are provided. Based on the previous stage, annotators should have learned visual and conceptual differences between classes; therefore, they are tasked to assign an appropriate class to each test image, by selecting either "class A" or "class B" for each of them. Contrary to the pre-learning stage, the option "I don't know" is not provided. 

After this stage, an accuracy score is extracted per user, based on their correct selections in the testing stage. We then extract an average accuracy per user, which we report in the main paper. Our average accuracy for the visually-informed experiment is 93.88\%, indicating that in most cases users are highly capable of recognizing the key concepts that separate the two given bird classes, grounding them with visual information. As for the blind experiment, the average testing accuracy is 89.28\%. Being rather close to the visually-informed accuracy percentage, we can safely assume that \textbf{concepts are more than adequate} towards teaching discriminative characteristics to humans, even if they lack association with purely visual information. Both visually-informed and blind accuracy scores clearly outperform the accuracy scores reported in CVE, demonstrating that conceptual explanations are more \textbf{meaningful} and \textbf{informative} to humans compared to pixel-level explanations.

\begin{wrapfigure}{r}{0cm}
    \centering
    \includegraphics[width=0.4\textwidth]{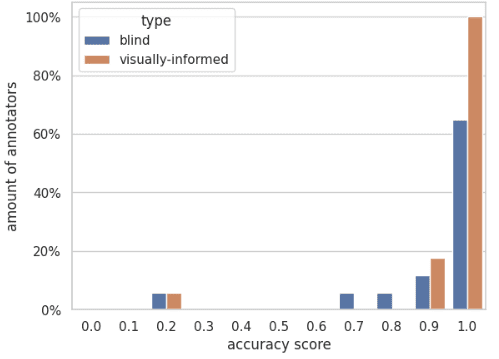}
    \caption{Distribution of test accuracy for machine teaching human evaluation experiments.}
    % \vskip -0.4in
    \label{fig:barplot_acc}
\end{wrapfigure}

\paragraph{Accuracy score distribution}
In Fig. \ref{fig:barplot_acc}, we present a more detailed analysis of the accuracy scores achieved by human subjects during the testing phase of the machine teaching experiment. It is apparent that scores peak at 0.9 and 1.0; thus, explanations produced by our method are highly human-interpretable and beneficial to perform classification. Comparison between 'visually-informed' and 'blind' results reveals that the decrease in test accuracy for the experiment without a visual aid is gradual.

\paragraph{Applicability of machine-teaching experiment}
% We select to run this 
The machine-teaching experiment is purposely run exclusively on the CUB dataset. 
To highlight the merits of the learning phase: annotators have no knowledge of bird species, therefore they can highly benefit from learning discriminative bird attributes, and then apply this new knowledge in the testing phase. For example, none of the annotators knows the difference between a Parakeet Auklet and a Least Auklet. Nevertheless, after the learning stage, they are able to recognize the basic discriminative attributes, which will help them classify instances of the test phase.
On the other hand, Visual Genome contains images of common everyday scenes, rendering a similar experiment rather redundant in such instances. For example, a human already knows key concepts that discriminate a kitchen from a bedroom, therefore the learning stage would be of no value, even if the scene labels are anonymized. We can view this scenario as an analog to data leakage. 

Moreover, there is always a possibility that some concepts can be misleading. In such cases, we expect visual classifiers to present a bias towards such concepts, while this is not the case for humans.
For example, a TV can be present in both kitchens and bedrooms. However, in a hypothetical scenario that selected bedroom images have TVs, but kitchen images do not, the graphs that serve as explanations would contain many "add TV" nodes. Therefore, a human expects to classify images containing TVs in one class, and images that do not contain TVs in the other (as a visual classifier would do if trained on such data). But when finally humans are presented with real test images, they will not be misled by the presence or the absence of TVs, but rather rely on their commonsense to perform classification. Thus, not only is the learning stage redundant, but the obvious existing bias "add TV" is not reflected in the final classification; in this case, the counterfactual explanation itself would be of no value to humans.

\section{Graph statistics}
\label{sec:graph_stats}
\begin{table*}[h!]
\begin{small}
\begin{center}
\centering
\caption{Statistics regarding graphs of different datasets used in the main paper.}
\label{tab:stats}
\vskip 0.15in
\begin{tabular}{cc  |c  |c  |c |c|c|c}
\toprule
&& VG-DENSE & VG-RANDOM & CUB 
&   D/P-SGG&D/P-CAPTION &SMARTY
\\
\midrule
\multirow{4}{4em}{Mean} &
 density & 0.20 & 0.06 & 0.04 
&   0.13&0.25 &0.23
\\
& edges & 9.04 & 8.77 & 27.52 
&   9.37&1.76 &4.40
\\
& nodes & 7.25 & 14.57 & 28.52 
&   9.73&3.20 &5.40
\\
& isolated nodes & 0.47 & 3.37 & 0 
&   0.32&0.90 &0
\\
\midrule
\multirow{4}{4em}{Max} &
density & 0.47 & 0.67 & 0.11
&   1.0&0.5 &0.33
\\
& edges & 36 & 27 & 53 
&   18&4 &15
\\
& nodes & 15 & 20 & 54 
&   18&5 &16
\\
& isolated nodes & 3 & 12 & 0 
&   3&4 &0
\\
\midrule
\multirow{4}{4em}{Min} &
 density & 0.14 & 0.01 & 0.02 
&   0.05&0.05 &0.06
\\
& edges & 5 & 5 & 8 
&   1&1 &2
\\
& nodes & 6 & 4 & 9 
&   2&2 &3
\\
& isolated nodes & 0 & 0 & 0 &   0&0 &0\\
\bottomrule
\end{tabular}
\end{center}
\end{small}
\end{table*}

\begin{wraptable}{r}{9.8cm}
\begin{small}
\begin{center}
\vskip -0.29in
\caption{Statistics regarding graphs of different datasets in the appendix.}
% \vskip 0.15in
\label{tab:stats2}
\vskip 0.15in
\centering
\begin{tabular}{cc  |P{2cm}|P{2cm}}
\toprule
&&AG &GQA 
\\
\midrule
\multirow{4}{4em}{Mean} &
 density &0.19
 &0.24
\\
& edges &13.23 &8.14
\\
& nodes &8.84 &6.66 
\\
& isolated nodes &1.17 &1.37 
\\
\midrule
\multirow{4}{4em}{Max} &
density &0.45
 &1.0 
\\
& edges &51 &20 
\\
& nodes &17 &12 
\\
& isolated nodes &2 &15 
\\
\midrule
\multirow{4}{4em}{Min} &
 density &0.1
 &0.13 
\\
& edges &4 &5 
\\
& nodes &5  &4 
\\
& isolated nodes &0 &0 \\
\bottomrule
\end{tabular}
\end{center}
\end{small}
\end{wraptable}
In Tab. \ref{tab:stats} we present some additional statistics regarding the graphs of the datasets used in our work (max and min details). VG-DENSE and VG-RANDOM contain 500 graphs each, CUB contains 422 graphs, D/P-SGG and D/P-CAPTION denote the web-crawled datasets with 259 graphs each and SMARTY denotes the COVID-19 classification dataset with 548 graphs. 
Table \ref{tab:stats2} contains additional statistics about datasets utilized only in the appendix. These are GQA \citep{hudson2019gqa} with 500 graphs mentioned in App. ref{sec:quant-additional} and Action Genome (AG) \citep{ji2020action} with 300 graphs mention in App. ref{sec:unannotated}.
The size and density of input data should be considered when viewing results in the experimental section.

\section{Experimental Settings}
\label{sec:setting-details}
In addition to details regarding resources used for the experimental setup mentioned in the main paper, we further report specific training configurations for GNN models. All presented results were achieved using single-layer GNNs of a dimension of 2048, built as explained in Sec. 3 of the main paper. For reproducibility purposes, we report that these models were optimized for a batch size of 32 and trained for 50 epochs, without the use of dropout. The employed optimizer was Adam without weight decay. The respective learning rate varied among GNN variants. To be precise, we used a learning rate of 0.04 for GCN and 0.02 for GAT and GIN. GAT and GIN also have model-specific hyperparameters - attention heads and the learnable parameter epsilon respectively. Best results were achieved by leveraging 8 attention heads and setting epsilon to non-learnable. 

Last but not least, an important hyperparameter of the GNN models is the number of training pairs, denoted as $p$. As explained, optimal models used $p \approx N/2$, which varies among datasets. Specifically, the parameter $p$ is set to 70K for datasets with 500 graphs, 50K for datasets with 422 graphs, and 25K for datasets with 300 graphs. However, we also conducted ablations on the number of training pairs, setting $p$ to values reported in Fig. 5 of the main paper. In those cases, we explored using 16\%, 40\%, and 80\% of the existent graph pairs, in addition to the "golden" ~50\%. 

Regarding graph kernels that were employed for comparison, we report that the Pyramid Match kernel was used with its default settings. The settings include leveraging labels, a histogram level $L$ of 4, and hypercube dimensions $d$ of 6.

The classifier in our CUB experiments (ResNet-50) was chosen in alignment to experiments performed in the works compared and recreated here. As for the choice of the Places365 instead of a pretrained ImageNet classifier, it was conscious. Despite the latte being potentially more widely recognized and researched, it is trained on the ImageNet dataset, which primarily consists of foreground objects. In Visual Genome, the majority of instances depict scenes, providing substantial background. Although some instances focus more on specific objects, they are still situated within a particular environment. In contrast, ImageNet classifiers face challenges with such inputs, as only about $3\%$ of the target classes in the corresponding dataset pertain to broader scenes. Classifiers for the rest of the datasets are explained in detail in the following sections.

The code for all experiments is provided within the zip file of the supplementary material, accompanied by comprehensive instructions.

\section{Quantitative Experiments}
\label{sec:quant}
\subsection{Graph Kernels}
Graph kernels are kernel functions used on graphs that measure similarity in polynomial time, providing an efficient and widely applicable alternative to GED. In the context of this paper, we experimented with several kernels from the GraKeL library \citep{siglidis2020grakel}, as a baseline measure for counterfactual retrieval. Our goal is to guarantee that our GNN framework outperforms such methods. We present results from the best-performing kernel Pyramid Match.

\paragraph{Pyramid Match (PM) kernel}
The PM \citep{grauman2007pyramid, nikolentzos2017matching} graph kernel operates by initially embedding each graph’s nodes in a d-dimensional vector space using the absolute eigenvectors of the largest eigenvalues of the adjacency matrix. The sets of graph vertices are compared by mapping the corresponding points in the d-dimensional hypercube to multi-resolution histograms, using a weighted histogram intersection function. The comparison process occurs in several levels, corresponding to different regions of the feature space with increasing size. The algorithm counts new matches at each level - i. e. points in the same region - and weights them according to the size of the level. The cells/regions double in size in each iteration of the algorithm.

This procedure is applicable to graphs with node/edge labels; thus, we cannot leverage GloVe embeddings for initialization. Matches exist only between points with the exact same label. The overall complexity of the algorithm is $O(ndL)$, which compared to other kernel methods is quite computationally expensive.

\subsection{Average GED}
\label{sec:avg-ged}
\input{tables-app/cub_avg_ged}
In addition to the average number of edits metric and the ranking metrics using the ground truth GED as the golden standard, we present the average GED of the top-1 counterfactual results. This supplementary measure serves to explicitly enhance comprehension of the significance of semantic context. Notably, within the main paper, our qualitative results illustrate scenarios where, despite an equal (or lower) number of edits, the GED can at times be higher. This divergence arises because edits are not uniformly weighted but rather based on their semantic similarity. 

\input{tables-app/refined_num_edits}

For the VG dataset, we present results comparing "Normal" and "Refined" outcomes. In this context, "Refined" denotes presenting averages exclusively when the two methods yield distinct counterfactuals. We adopted this approach due to the observation that $75\%$ of CEs for VG-DENSE and $73\%$ for VG-RANDOM were identical between methods, creating an impression of increased result proximity. To provide a comprehensive view, we furnish more refined average number of edits results in Table \ref{tab:n_edits-vg-refined}. Notably, for the CUB dataset, such an analysis is unnecessary; nonetheless, we include the average top-1 GED in Table \ref{tab:avg-ged-cub}.

\subsection{Additional Datasets}
\label{sec:quant-additional}

\begin{table*}[h!]

% \vskip -0.1in
\begin{center}
\hspace{-0.4em}
\caption{Ranking results on GQA for different graph models.}
\label{tab:hit-percentage-gqa}
\vskip 0.15in
\begin{tabular}{c|p{0.5cm}p{0.5cm}p{0.6cm}|p{0.5cm}p{0.5cm}p{0.6cm}|p{0.5cm}p{0.5cm}p{0.6cm}|p{0.5cm}p{0.5cm}p{0.6cm}}
\toprule
& \multicolumn{3}{c}{\small P@k  $\uparrow$} & \multicolumn{3}{c}{\small NDCG@k $\uparrow$} & \multicolumn{3}{c}{\small P@k (binary) $\uparrow$} & \multicolumn{3}{c}{\small NDCG@k (binary) $\uparrow$}\\  
\cline{2-13}
 & k=1 & k=2 & k=4 & k=1 & k=2 & k=4  & k=1 & k=2 & k=4 & k=1 & k=2 & k=4 \\
\midrule
\small PM & \small 0.06 &  \small 0.10 &  \small 0.05 &  \small 0.65 &  \small 0.65 &  \small 0.68  &  \small 0.10 &  \small 0.15 &  \small 0.20 &  \small 0.17 &  \small 0.22 &  \small 0.31  \\

\small GIN-70K & \small 0.16 &  \small 0.24 &  \small 0.29  &  \small 0.70 &  \small 0.70 &  \small 0.72 &  \small 0.16 &  \small 0.27 &  \small 0.39  &  \small 0.20 &  \small 0.25 &  \small 0.34 \\
\small GAT-70K & \small 0.13 &  \small 0.17 &  \small 0.22  &  \small 0.66 &  \small 0.68 &  \small 0.69 &  \small 0.13 &  \small 0.21 &  \small 0.30  &  \small 0.18 &  \small 0.24 &  \small 0.32 \\
\small GCN-70K & \small \textbf{0.19} &  \small \textbf{0.29} &  \small \textbf{0.34} &  \small \textbf{0.73} &  \small \textbf{0.73} &  \small \textbf{0.74} &  \small \textbf{0.19} &  \small \textbf{0.33} &  \small \textbf{0.48} &  \small \textbf{0.22} &  \small \textbf{0.28} &  \small \textbf{0.36}\\
\bottomrule
\end{tabular}
\end{center}
% \vskip -0.1in
\end{table*}

\paragraph{Ranking results for GQA}
\begin{wrapfigure}{r}{0cm}
    % \vskip -0.17in
    \centering
    \includegraphics[width = 0.4\textwidth]{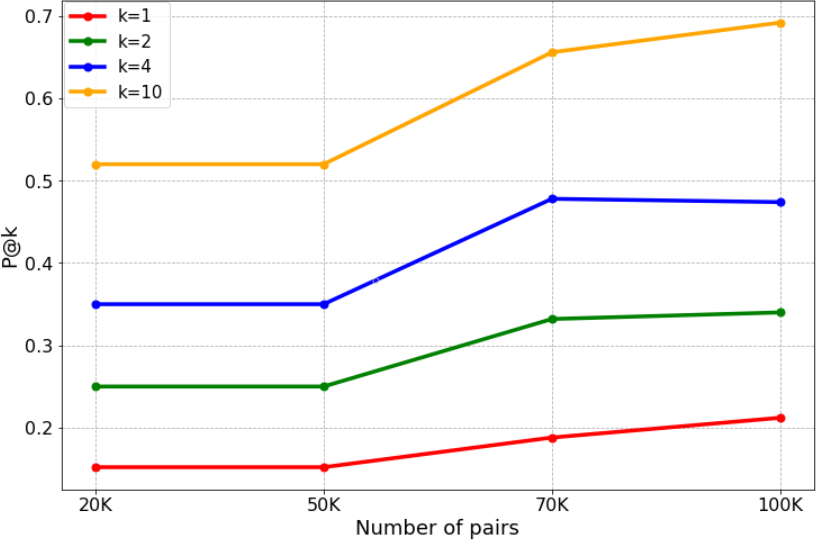}
    \caption{Comparison of the GCN performance measured in P@k for different number of training pairs $p$ for GQA.}
    \label{fig:gqa-pairs}
    % \vskip -0.5in
\end{wrapfigure}

The analysis performed on Visual Genome (VG) is extended on the GQA dataset \citep{hudson2019gqa}. In fact, GQA comprises a variant of VG focusing on compositional question-answering involving real-world scenes. Since GQA images and accompanying scene graphs are very similar to the ones involved in our VG analysis, the obtained results verify the findings reported for VG without offering other novel insights.
In Tab. \ref{tab:hit-percentage-gqa} we present per-model results for 70K training pairs. GCN remains the most powerful architecture compared to the other ones, an observation validating the findings reported for the rest of the datasets.

\paragraph{Performance-complexity trade-off for GQA}
In Fig. \ref{fig:gqa-pairs} we present the performance analysis for different numbers of training pairs $p$ on the GQA dataset, focusing on our best-performing model (GCN). Once again, $N/2 \sim$ 70K pairs are adequate for learning proper representations of scene graphs, validating our initial claim that GED does not have to be computed for more than $N/2$ graph pairs to obtain a satisfactory approximation.

\paragraph{Ranking results for Action Genome} are presented in Tab. \ref{tab:hit-percentage-ag}, while \textbf{number of edits for Action Genome} is presented in Tab. \ref{tab:n_edits-ag}, both for $N/2$ = 25K. 

\begin{table*}[h!]

% \vskip -0.1in
\begin{center}
% \hspace{-0.4em}
\caption{Ranking results on AG.}
\label{tab:hit-percentage-ag}
\vskip 0.1in
\begin{tabular}
{c|p{0.5cm}p{0.5cm}p{0.6cm}|p{0.5cm}p{0.5cm}p{0.6cm}|p{0.5cm}p{0.5cm}p{0.6cm}|p{0.5cm}p{0.5cm}p{0.6cm}}
\toprule
& \multicolumn{3}{c}{\small P@k  $\uparrow$} & \multicolumn{3}{c}{\small NDCG@k $\uparrow$} & \multicolumn{3}{c}{\small P@k (binary) $\uparrow$} & \multicolumn{3}{c}{\small NDCG@k (binary) $\uparrow$}\\  
\cline{2-13}
 & k=1 & k=2 & k=4 & k=1 & k=2 & k=4  & k=1 & k=2 & k=4 & k=1 & k=2 & k=4 \\
\midrule
\small GCN-25K & \small 0.17  &  \small 0.21 &  \small 0.27 &  \small 0.70 &  \small 0.70 &  \small  0.72 &  \small 0.17 &  \small 0.26 &  \small 0.41 &  \small 0.21 &  \small 0.27 &  \small 0.35  \\
\bottomrule
\end{tabular}

\end{center}
% \vskip -0.1in
\end{table*}

% \paragraph{Number of edits for Action Genome} for $N/2$ = 25K are presented in Table \ref{tab:n_edits-ag}.
\begin{table}[h!]
%\vskip -0.12in
\centering
\caption{Average number of node, edge \& total edits on AG.
% \textbf{Bold} denotes best results.
}
%\vskip -0.1in
\label{tab:n_edits-ag}
\vskip 0.15in
\begin{tabular}{c|ccc} 
\toprule
% &  \multicolumn{3}{c}{\small AG}  \\
% \cline{2-4}
& \small Node$\downarrow$ & \small Edge$\downarrow$ & \small Total$\downarrow$\\
\midrule
\small GCN-25K  &  \small 4.87  &  \small 7.99 &  \small 12.86 \\
\bottomrule
\end{tabular} 
% \vskip -0.1in
\end{table}

\subsection{Global edits on CUB}
\label{sec:global-cub}
\begin{figure}[h!]
    \centering
\begin{subfigure}{.45\textwidth}
        \centering
        \includegraphics[width=0.7\linewidth]{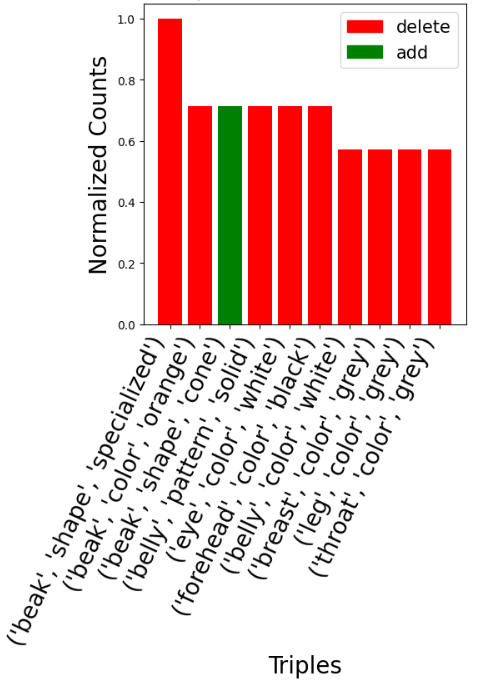}
        \caption{}
        \label{fig:cub-triples}
    \end{subfigure}\hfill
    \hspace{1cm} % Adjust the vertical spacing between subfigures
    %\vskip -2in
    \begin{subfigure}{.45\textwidth}
        \centering        \includegraphics[width=0.7\linewidth]{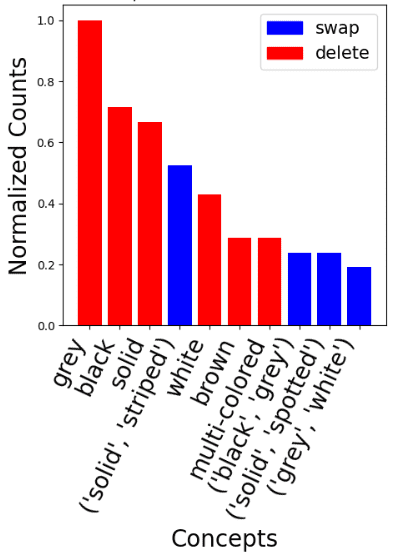}
        % \caption{Concept edits (\textcolor{ForestGreen}{insertions}, \textcolor{red}{deletions}, \textcolor{blue}{substitutions}) to perform Parakeet Auklet $\rightarrow$ Least Auklet transition.}
        \caption{}
        \label{fig:cub-concepts}
    \end{subfigure}
    \caption{Triple and concept edits (\textcolor{ForestGreen}{insertions}, \textcolor{red}{deletions}, \textcolor{blue}{substitutions}) to perform Parakeet Auklet $\rightarrow$ Least Auklet transition.}
    \label{fig:combined-figures}
\end{figure}

By aggregating edits from each image participating in the dataset, we can extract \textit{global edits}: they describe what needs to be changed in total to explain the transition from one class to the other. These edits are more meaningful in the form of graph triples, but we can also provide concept or relationship edits. In Figure \ref{fig:cub-triples}, we provide the triple edits to explain the Parakeet Auklet $\rightarrow$ Least Auklet counterfactual transition.
Similarly, in Figure \ref{fig:cub-concepts} we present global edits for concepts appearing on CUB images. The results align with human perception.

\section{Qualitative analysis}
\label{sec:qual}
\subsection{Counterfactual graph geometry on CUB}
\begin{figure}[h!]
     \centering
     \begin{subfigure}[b]{0.32\textwidth}
         \centering
         \includegraphics[width=2.5cm, height = 3.5cm]{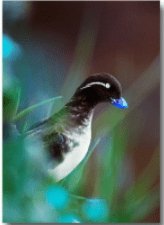}
         \caption{Query image}
         \label{fig:gt-fig}
     \end{subfigure}
     \hfill
     \begin{subfigure}[b]{0.33\textwidth}
         \centering
         \includegraphics[width=2.5cm, height = 3.5cm]{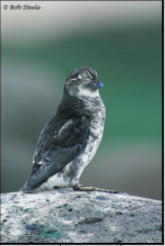}
         \caption{Top-1 retrieved by CVE %\cite{vandenhende2022making}
         }
         \label{fig:fb-fig}
     \end{subfigure}
     \hfill
     \begin{subfigure}[b]{0.32\textwidth}
         \centering
         \includegraphics[width=2.5cm, height = 3.5cm]{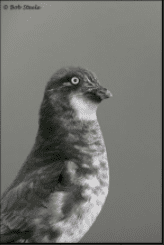}
         \caption{Top-1 retrieved (ours)}
         \label{fig:ours-fig}
     \end{subfigure}
        \caption{A counterfactual explanation example.}
        \label{fig:example-fig}
\end{figure}

Our framework is capable of retrieving counterfactual graphs that not only respect node and edge semantics, but also graph geometry. 
This observation corresponds to more accurate retrieval capabilities that focus on semantic information regarding bird species without being significantly distracted from irrelevant characteristics such as the background. This can be an encouraging characteristic of our counterfactuals towards more robust explanations, even though this aspect is not analyzed in the current paper. First, we present a qualitative example of this claim. In Figure \ref{fig:example-fig}, we search for the most similar image to \ref{fig:gt-fig} using the method of CVE and ours.

\begin{figure}[h!]
\centering
\begin{subfigure}[b]{0.3\textwidth}
\centering
\includegraphics[width=6.5cm, height = 6.5cm] {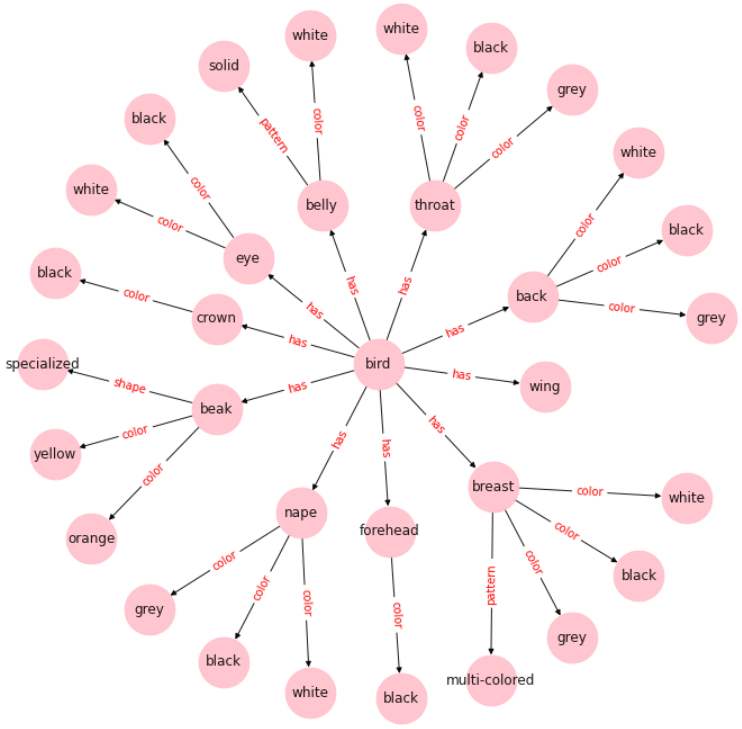}
\caption{Graph of class Parakeet Auklet corresponding to query image of Fig. \ref{fig:gt-fig}.}
\label{fig:gt}
\end{subfigure} 
\vskip\baselineskip
\begin{subfigure}[b]{0.3\textwidth}
\centering
\includegraphics[width=6.5cm, height = 6.5cm] {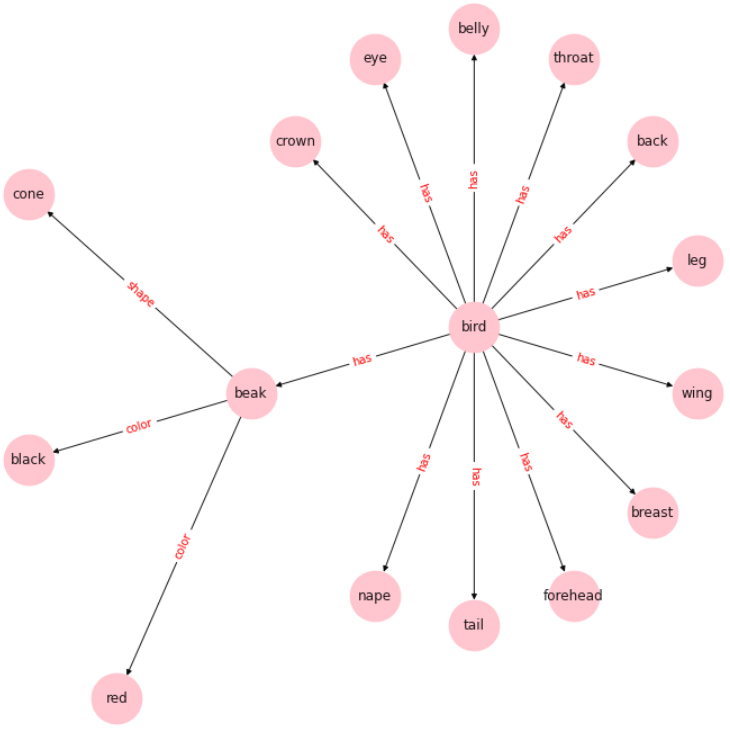}
\caption{Counterfactual graph of target class Least Auklet corresponding to Fig. \ref{fig:fb-fig} (as retrieved by CVE).}
\label{fig:fb}
\end{subfigure}
\hskip 1.2in
\begin{subfigure}[b]{0.3\textwidth}
\centering
\includegraphics[width=6.5cm, height = 6.5cm] {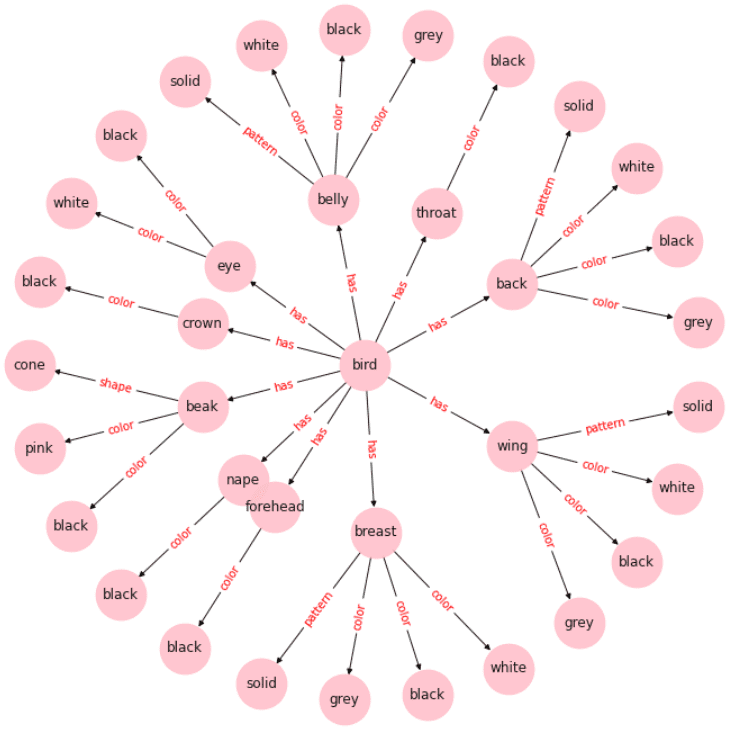}
\caption{Counterfactual graph of target class Least Auklet for Fig. \ref{fig:ours-fig} (as retrieved by our GCN-70K).}
\label{fig:ours}
\end{subfigure}
\caption{Example of scene graph structures of counterfactual graphs for Parakeet Auklet $\rightarrow $ Least Auklet class transition.}
\label{fig:example}
\end{figure}

Apparently, both counterfactual images are visually similar, as appearing in Fig. \ref{fig:fb-fig} and \ref{fig:ours-fig}. 
% According to human perception, it is hard to discriminate which method was more capable of capturing semantically significant features.
However, the representation power of scene graphs becomes evident in this case. 
In Fig.  \ref{fig:example} we present the scene graphs corresponding one-to-one to the images of Fig. \ref{fig:example-fig}.
The most similar graphs of \ref{fig:gt} correspond to the graph of \ref{fig:fb} according to CVE and \ref{fig:ours} according to our approach. It is evident that our approach can successfully retrieve graphs that \textbf{better respect the geometry} of the source image scene graph. Another observation is that our approach manages to retrieve an image without the concepts 'leg' or 'tail' which is more accurate compared to the source. Therefore, structural similarity leads to better semantic consistency.

\subsection{Graphs of Visual Genome}
\label{sec:graphs_vg}
\begin{figure}[h!]
\centering
     \begin{subfigure}[b]{0.32\textwidth}
         \centering
        \includegraphics[width=4.5cm, height = 4.5cm]{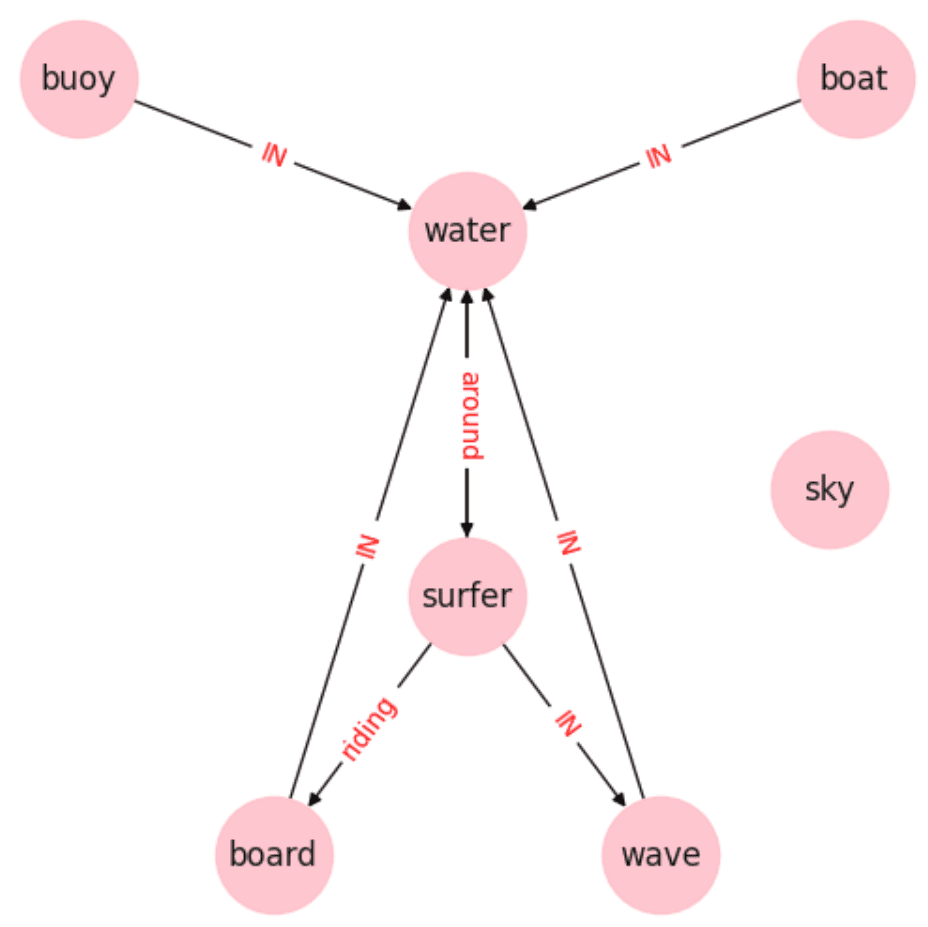}
         %\caption*{}
     \end{subfigure}
     \hfill
     \begin{subfigure}[b]{0.32\textwidth}
         \centering
        \includegraphics[width=4.5cm, height = 4.5cm]{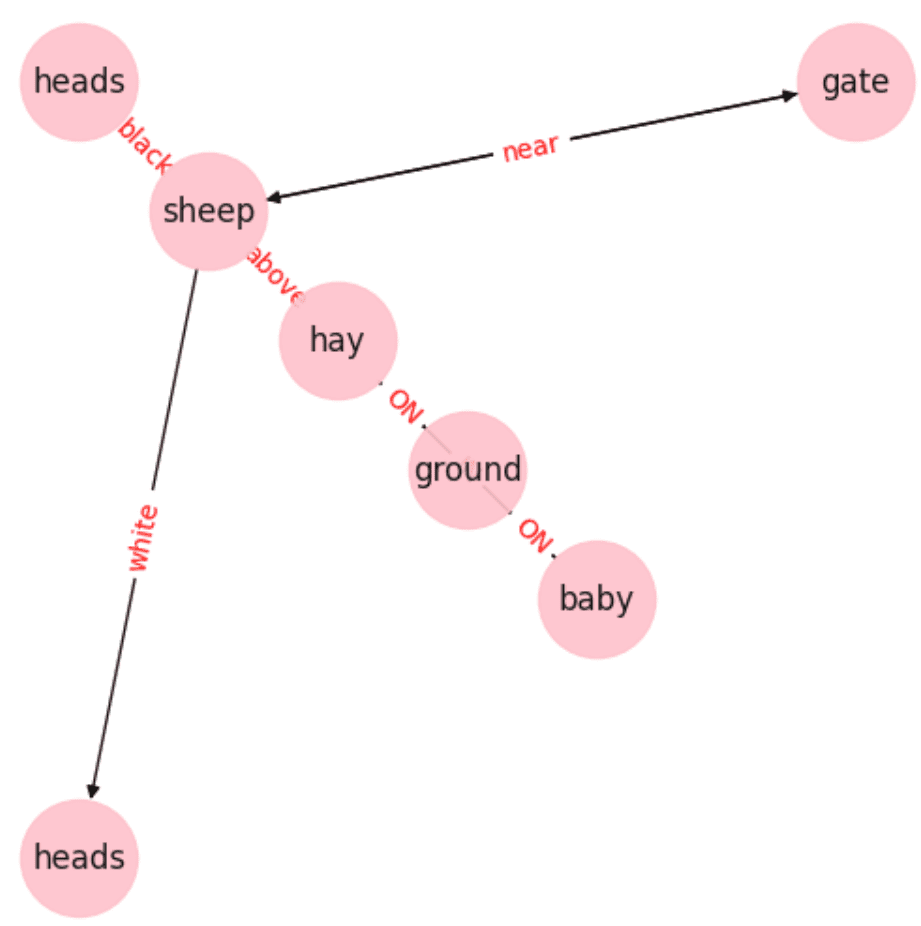}
         %\caption*{}
     \end{subfigure}
     \hfill
     \begin{subfigure}[b]{0.32\textwidth}
         \centering
        \includegraphics[width=4.5cm, height = 4.5cm]{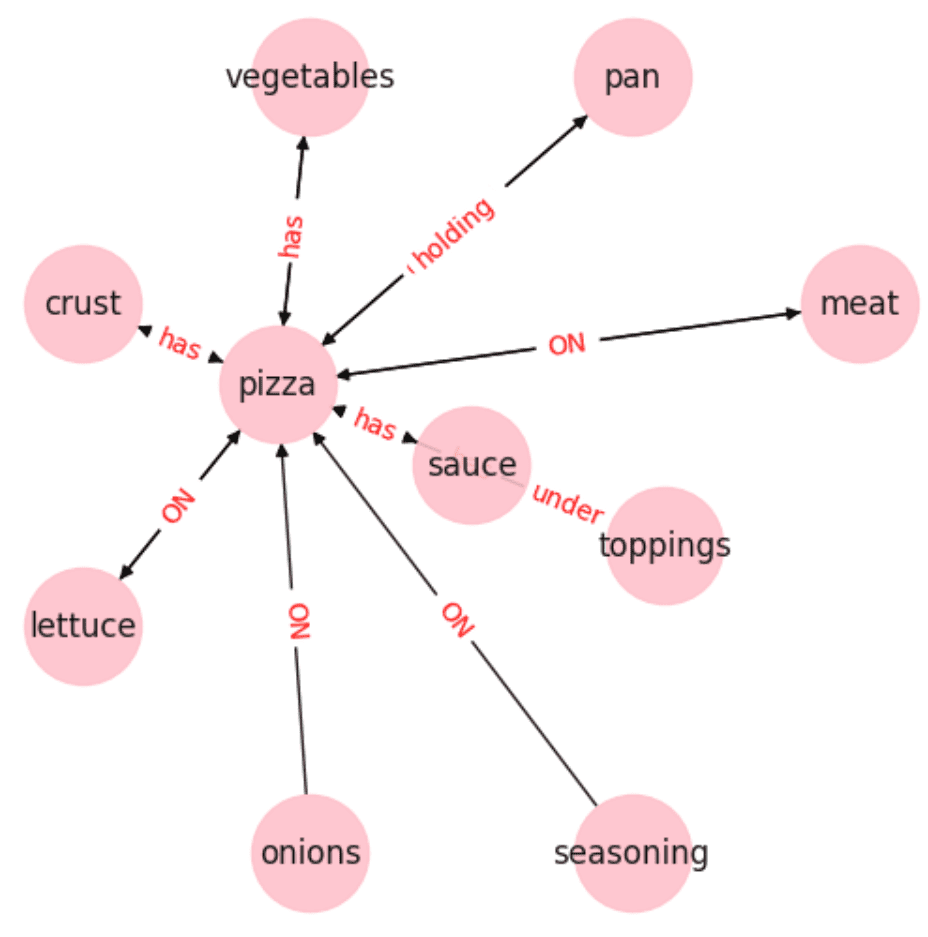}
         %\caption*{}
     \end{subfigure}
\par (a) Source Graphs
\label{fig:source-vg-dense-g}
\vskip\baselineskip
     \begin{subfigure}[b]{0.32\textwidth}
         \centering
        \includegraphics[width=4.5cm, height = 4.5cm]{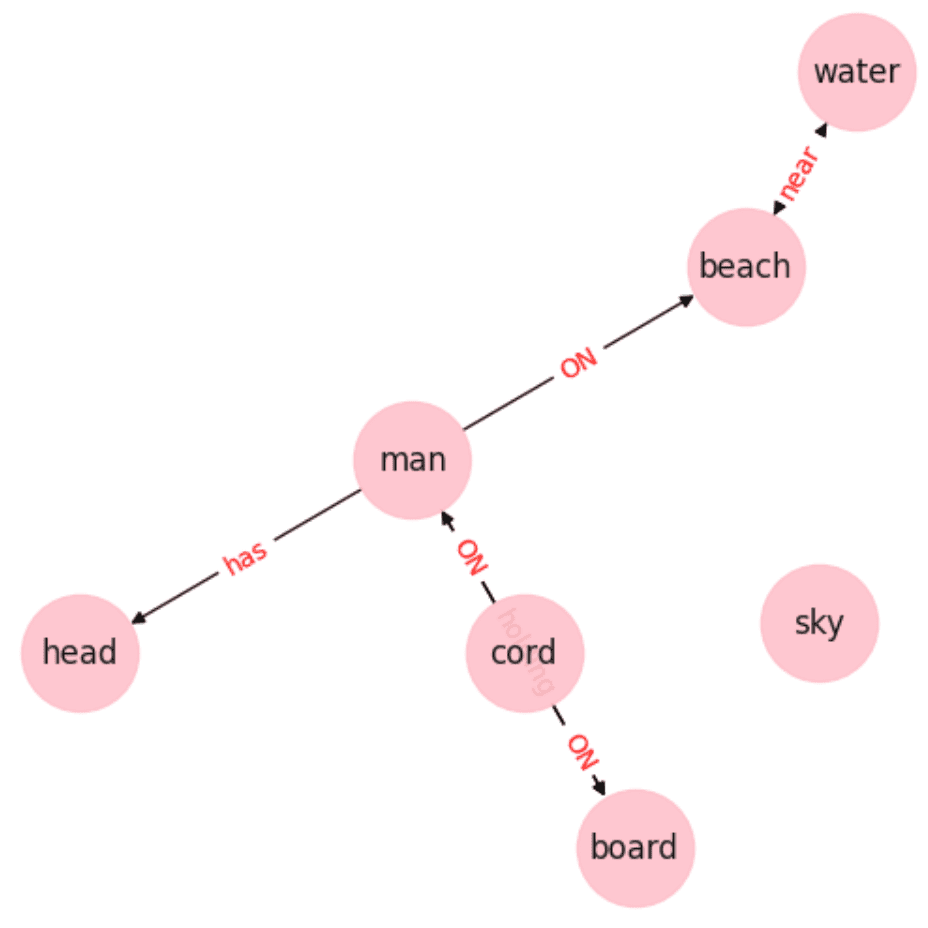}
     \end{subfigure}
     \hfill
     \begin{subfigure}[b]{0.32\textwidth}
         \centering
        \includegraphics[width=4.5cm, height = 4.5cm]{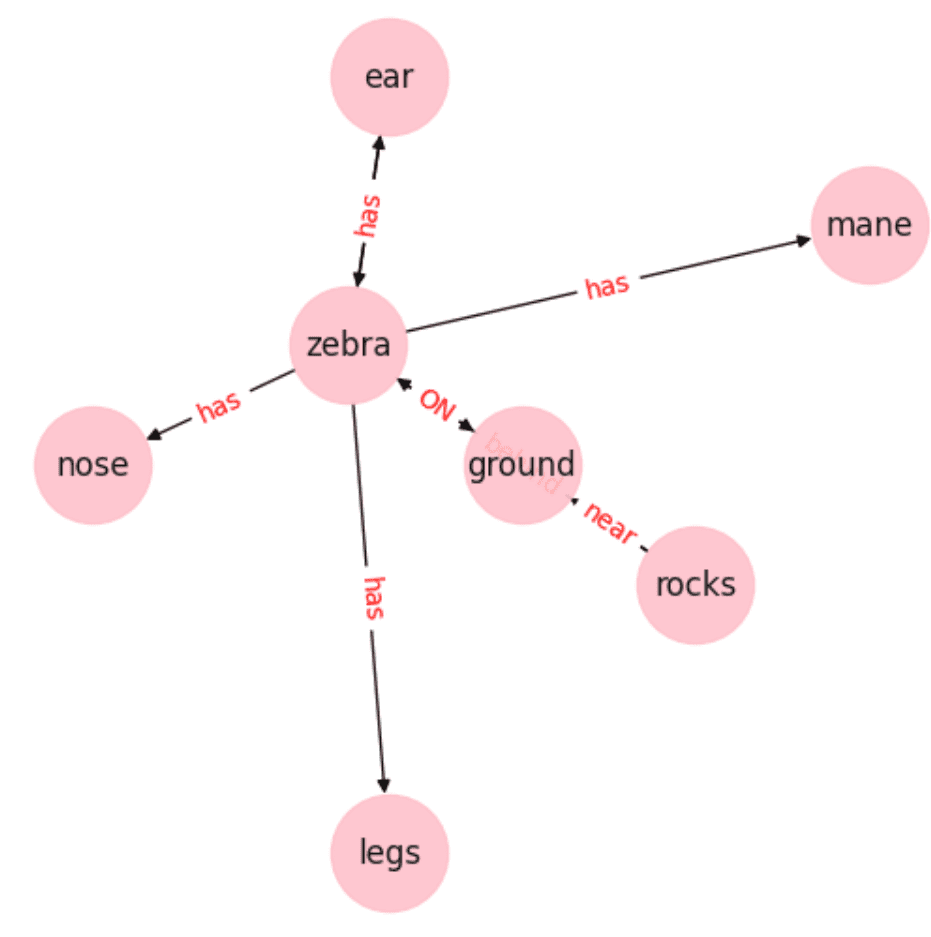}
     \end{subfigure}
     \hfill
     \begin{subfigure}[b]{0.32\textwidth}
         \centering
        \includegraphics[width=4.5cm, height = 4.5cm]{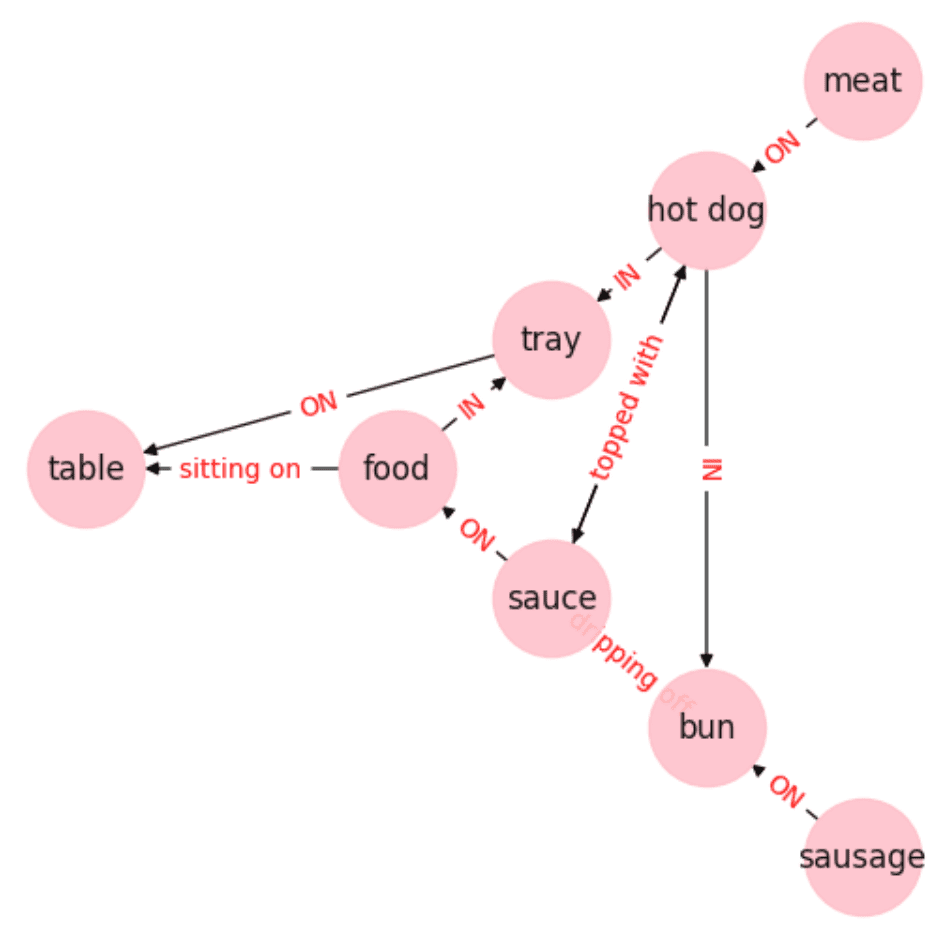}
     \end{subfigure}
\par (b) Counterfactual graphs using the method of SC
\label{fig:cece-vg-dense-g}
\vskip\baselineskip
     \begin{subfigure}[b]{0.32\textwidth}
         \centering
         \includegraphics[width=4.5cm, height = 4.5cm]{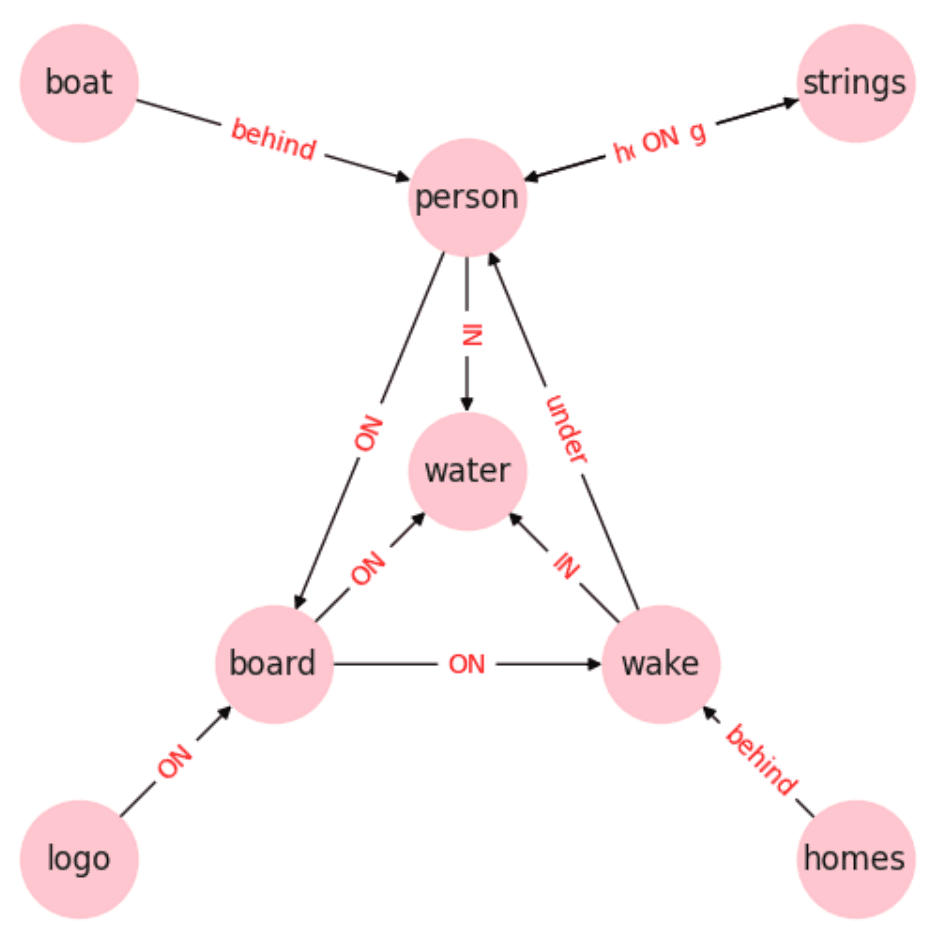}
     \end{subfigure}
     \hfill
     \begin{subfigure}[b]{0.33\textwidth}
         \centering
         \includegraphics[width=4.5cm, height = 4.5cm]{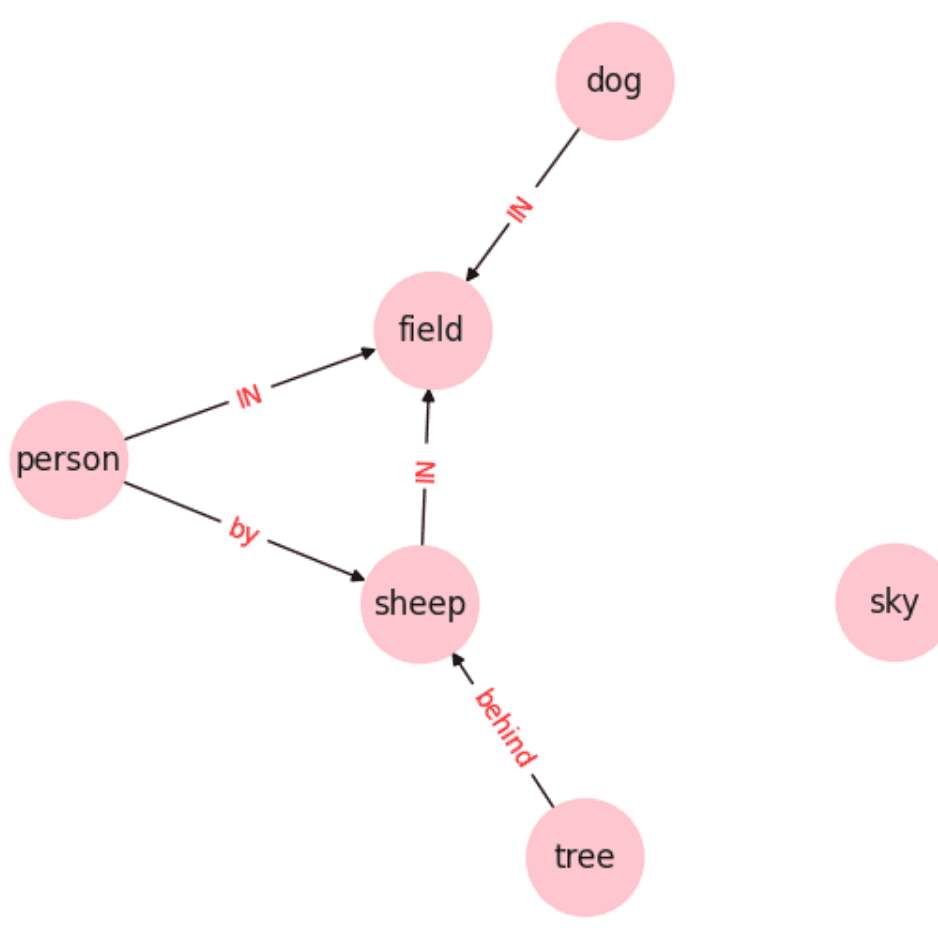}
     \end{subfigure}
     \hfill
     \begin{subfigure}[b]{0.32\textwidth}
         \centering
         \includegraphics[width=4.5cm, height = 4.5cm]{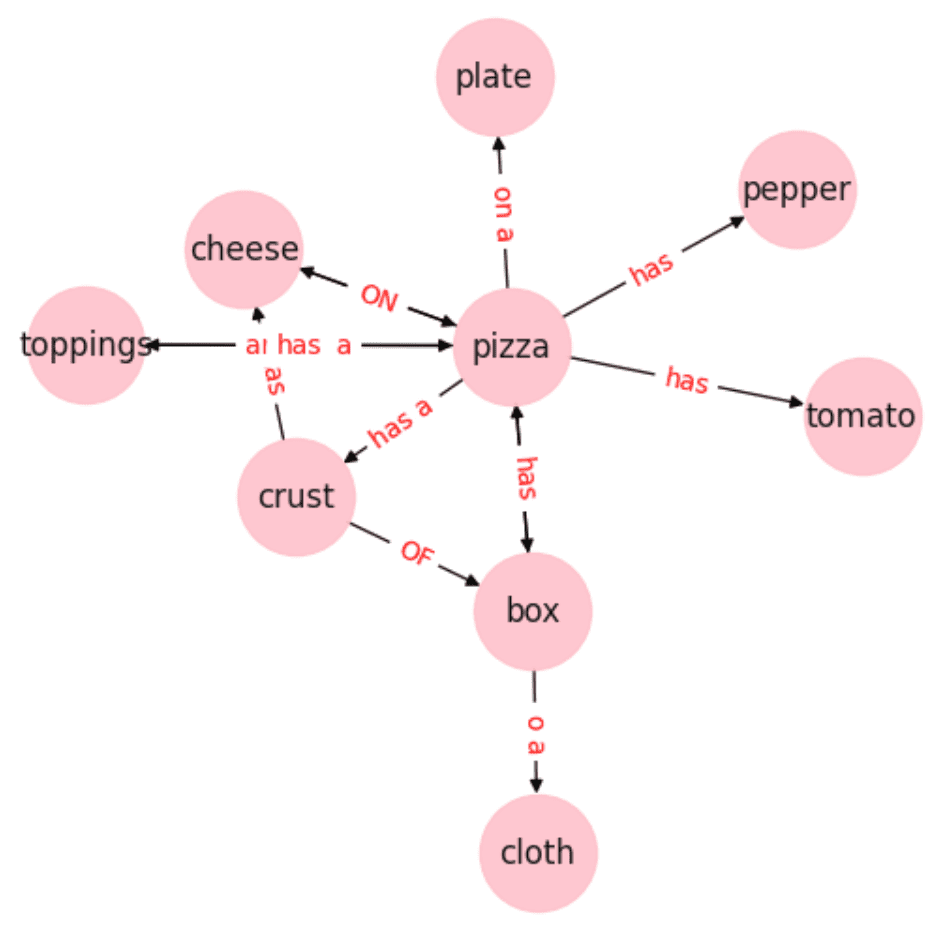}
     \end{subfigure}
\par (d) Counterfactual graphs (ours)
\label{fig:gnn-vg-dense-g}

\caption[]{
Qualitative Results on graphs for counterfactuals presented in Fig. 4 of the main paper for VG-DENSE.}
\label{fig:vg-dense-g}
\end{figure}

\begin{figure}[h!]
\centering
     \begin{subfigure}[b]{0.32\textwidth}
         \centering
        \includegraphics[width=4.5cm, height = 4.5cm]{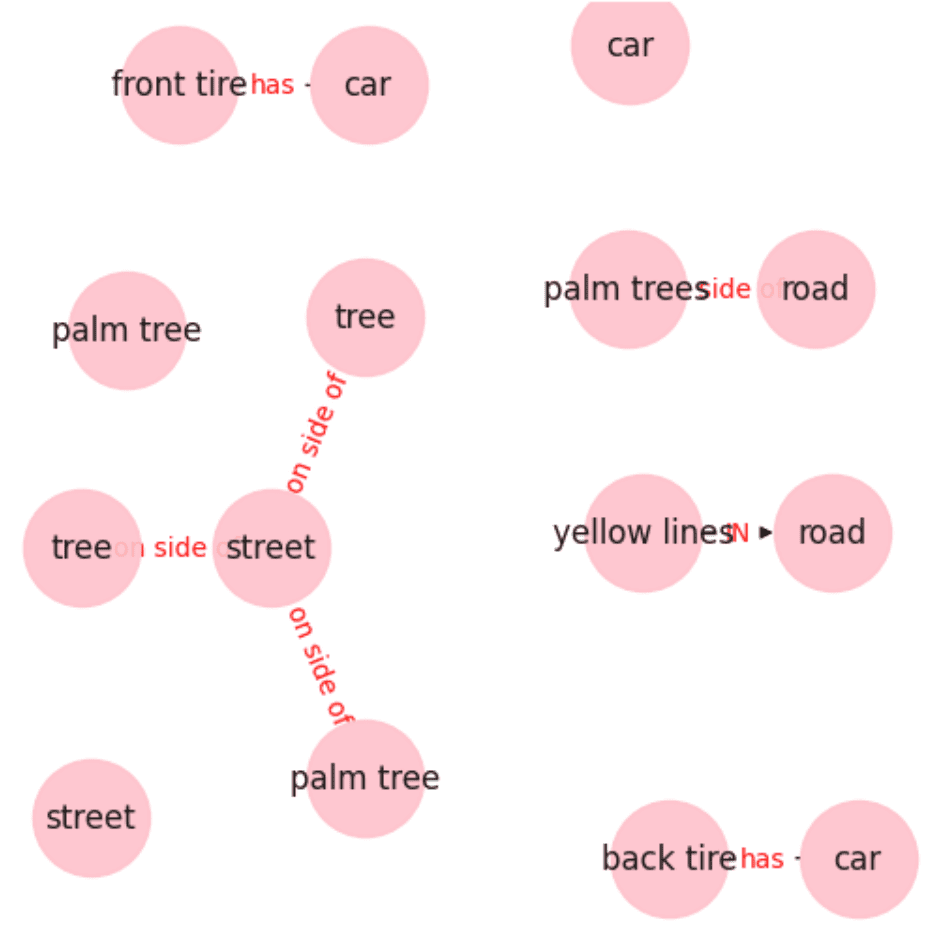}
         %\caption*{}
     \end{subfigure}
     \hfill
     \begin{subfigure}[b]{0.32\textwidth}
         \centering
        \includegraphics[width=4.5cm, height = 4.5cm]{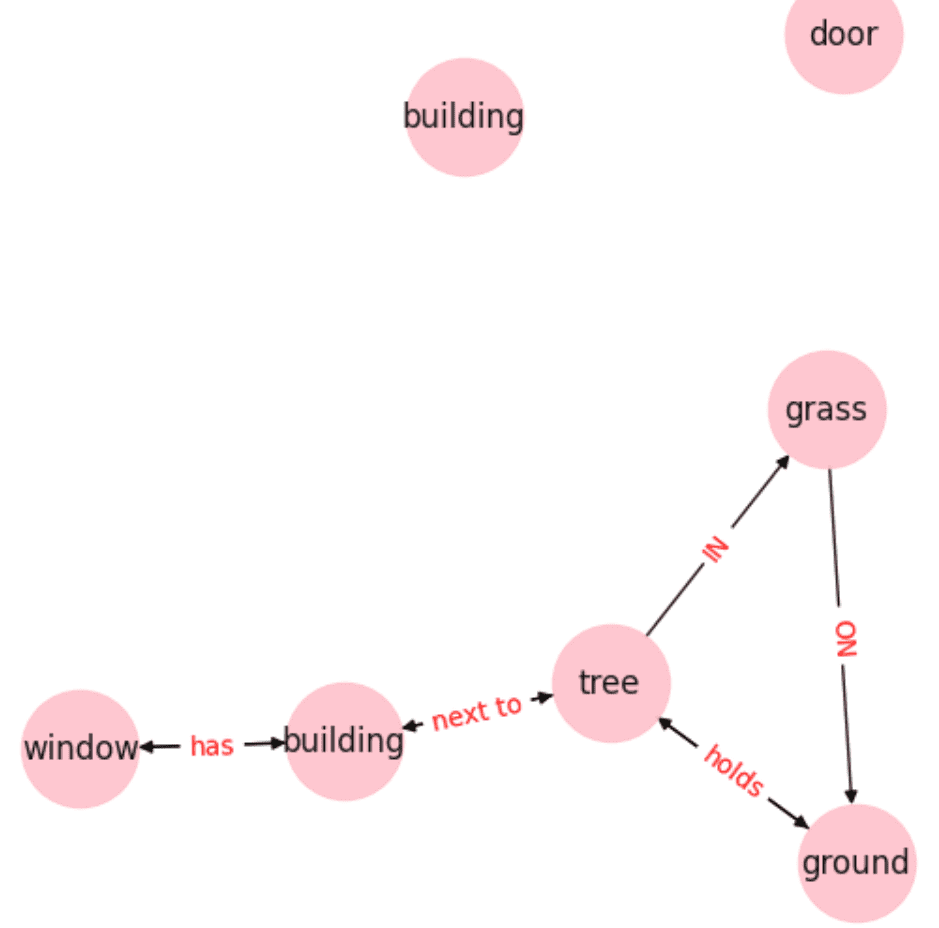}
         %\caption*{}
     \end{subfigure}
     \hfill
     \begin{subfigure}[b]{0.32\textwidth}
         \centering
        \includegraphics[width=4.5cm, height = 4.5cm]{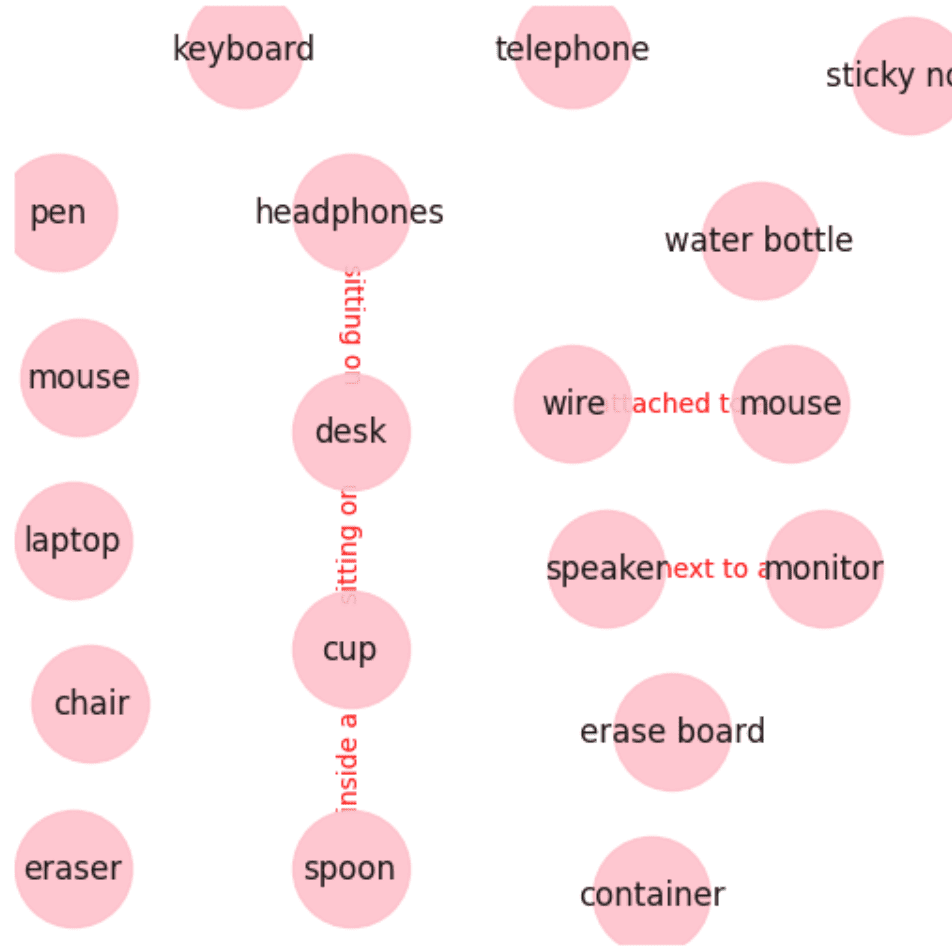}
         %\caption*{}
     \end{subfigure}
\par (a) Source Graph
\label{fig:source-vg-random-g}
\vskip\baselineskip
     \begin{subfigure}[b]{0.32\textwidth}
         \centering
        \includegraphics[width=4.5cm, height = 4.5cm]{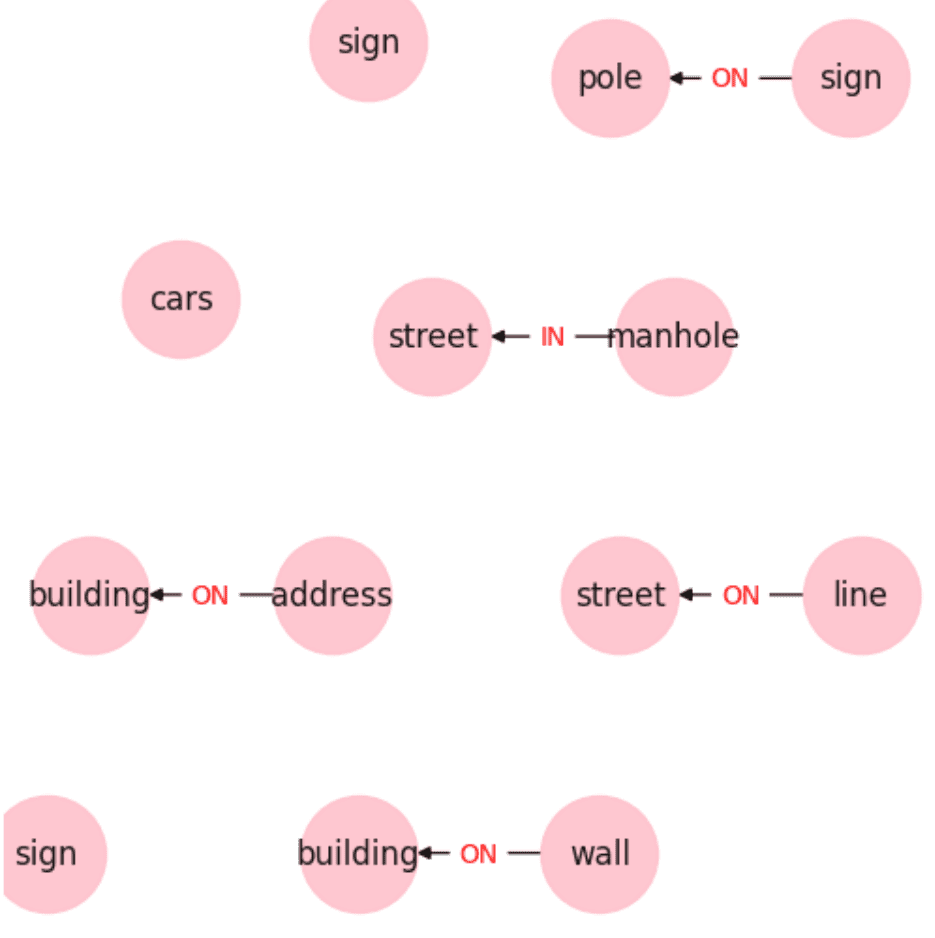}
     \end{subfigure}
     \hfill
     \begin{subfigure}[b]{0.32\textwidth}
         \centering
        \includegraphics[width=4.5cm, height = 4.5cm]{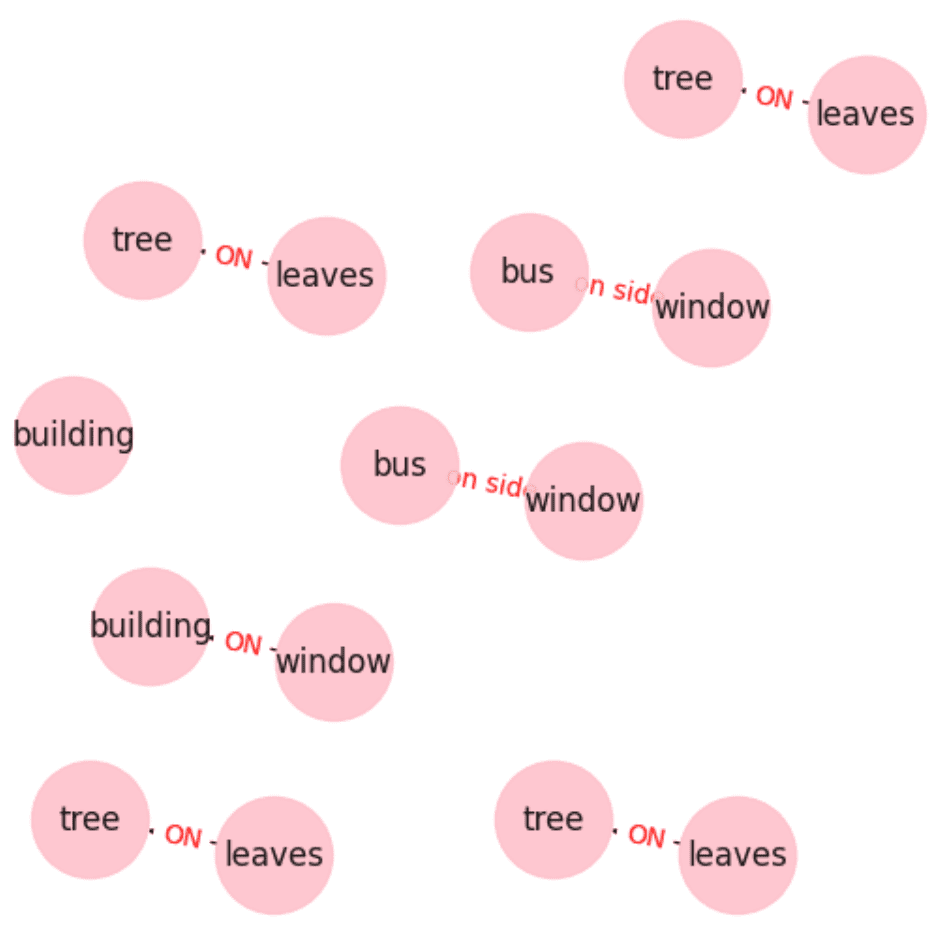}
     \end{subfigure}
     \hfill
     \begin{subfigure}[b]{0.32\textwidth}
         \centering
        \includegraphics[width=5cm, height = 4cm]{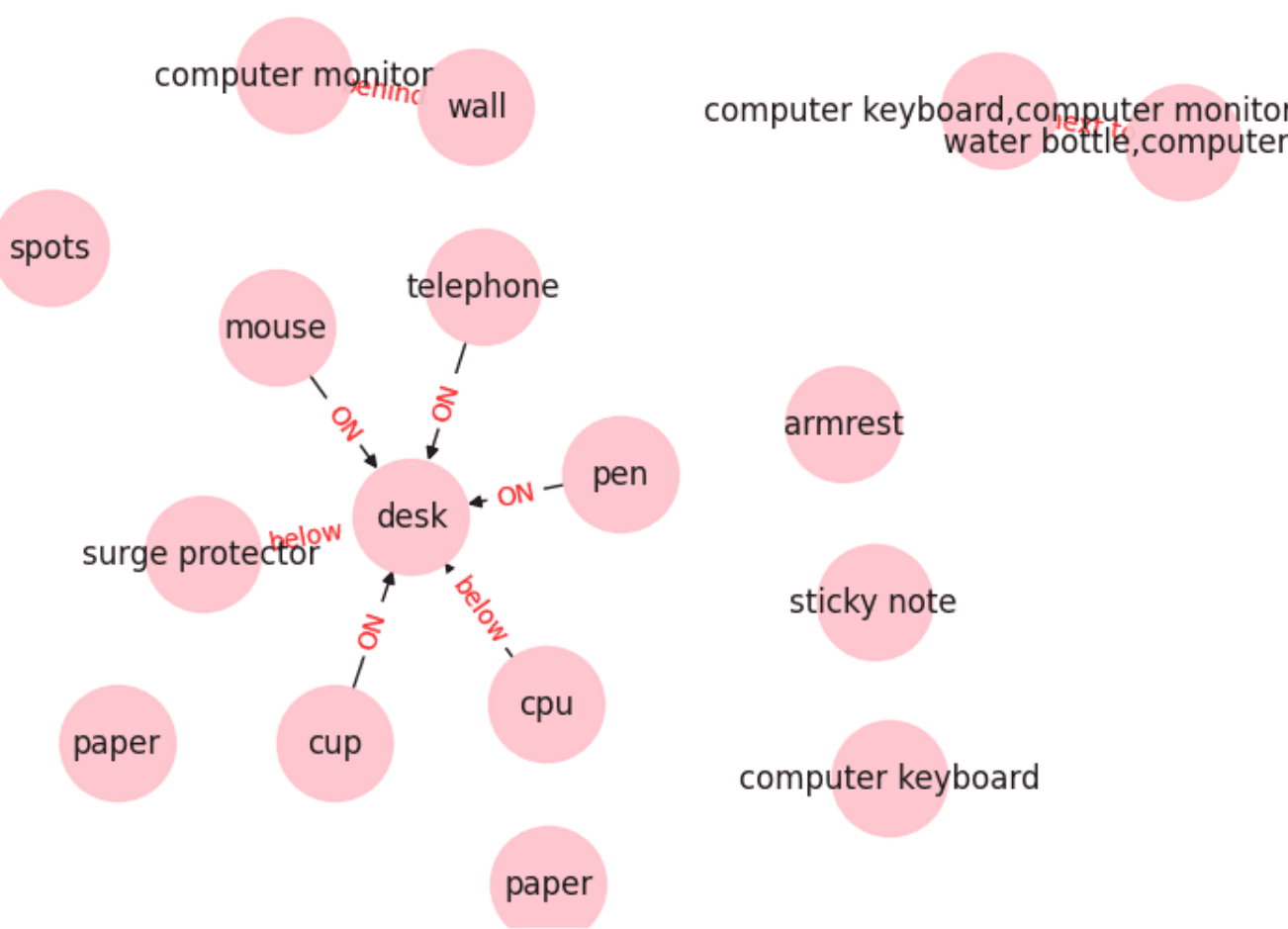}
     \end{subfigure}
\par (b) Counterfactual (SC)
\label{fig:cece-vg-random-g}
\vskip\baselineskip
     \begin{subfigure}[b]{0.32\textwidth}
         \centering
         \includegraphics[width=4cm, height = 4cm]{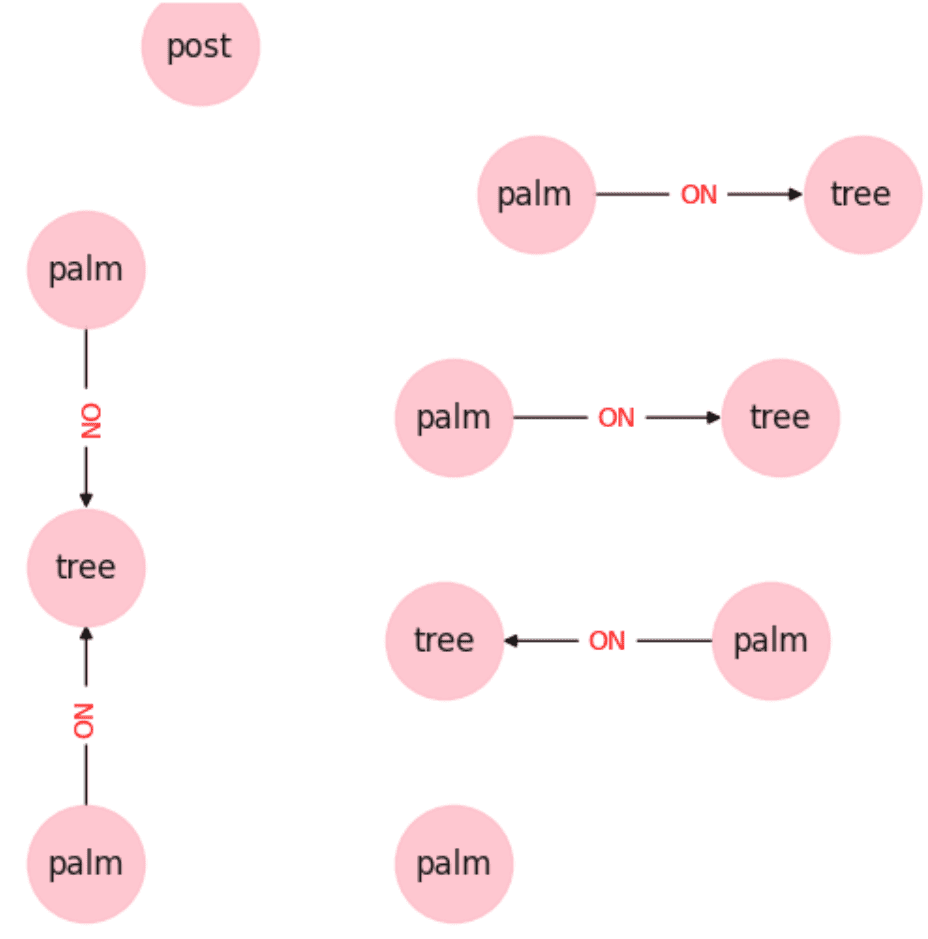}
     \end{subfigure}
     \hfill
     \begin{subfigure}[b]{0.33\textwidth}
         \centering
         \includegraphics[width=4.5cm, height = 4.5cm]{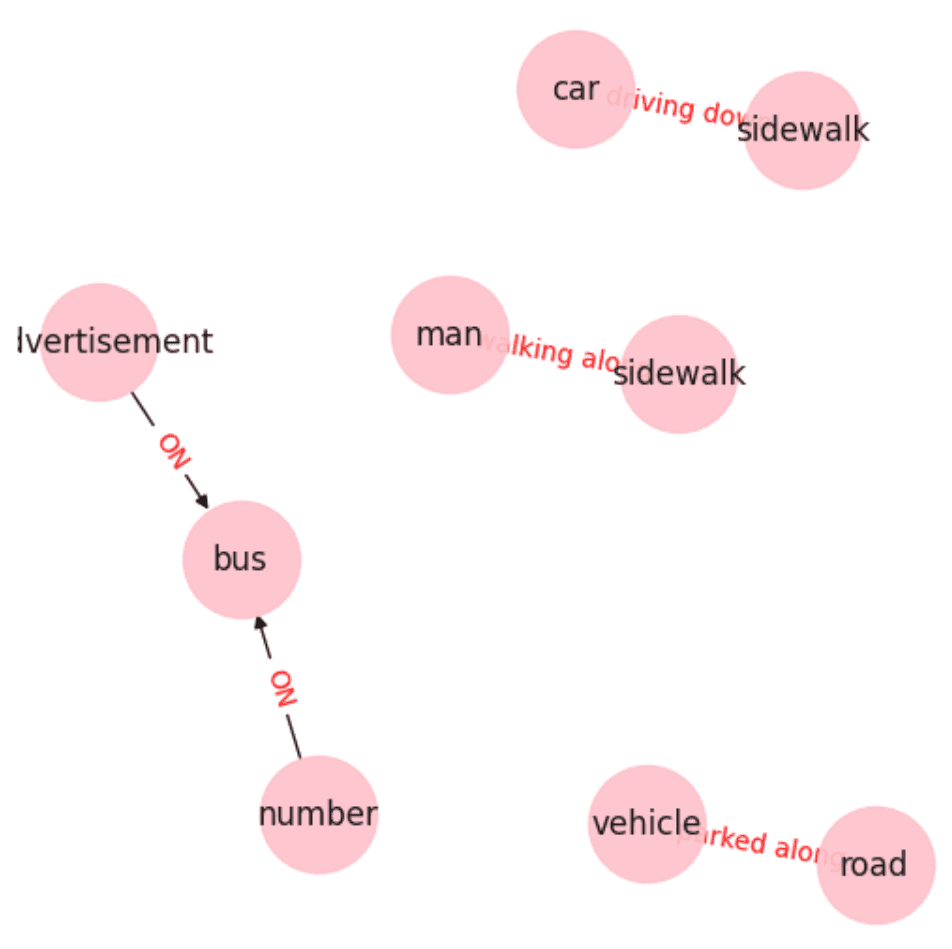}
     \end{subfigure}
     \hfill
     \begin{subfigure}[b]{0.32\textwidth}
         \centering
         \includegraphics[width=4.5cm, height = 4.5cm]{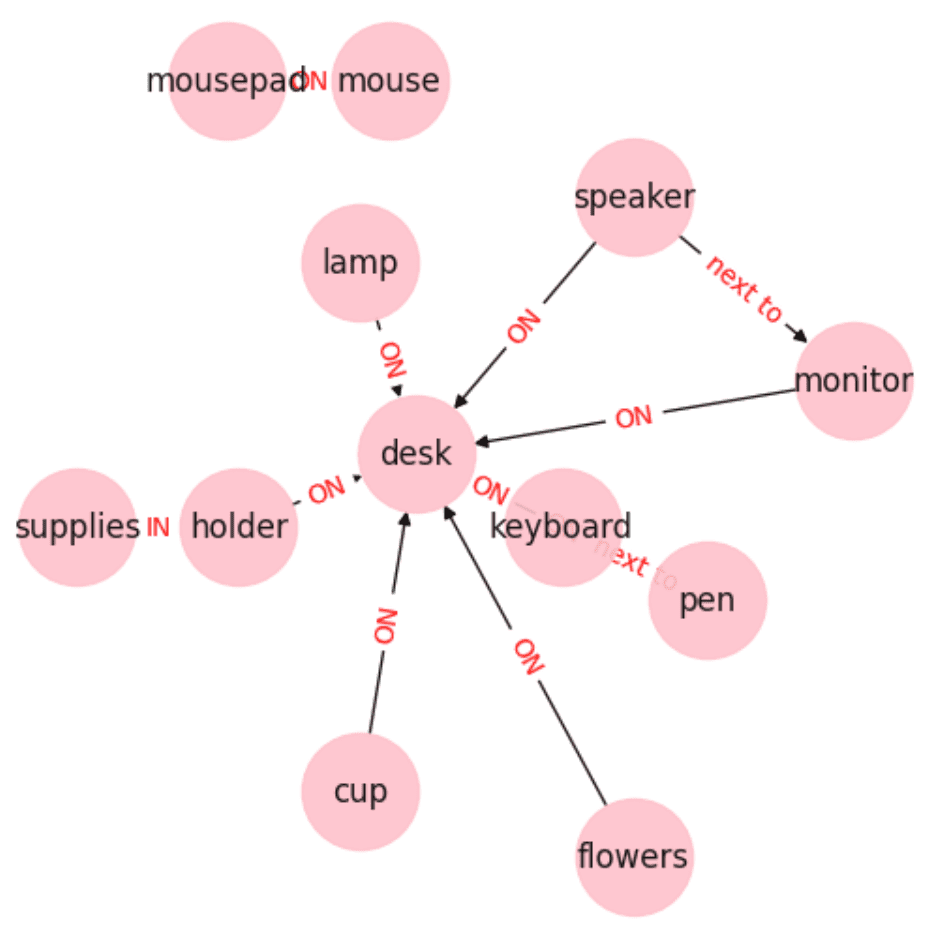}
     \end{subfigure}
\par (d) Counterfactual (ours)
\label{fig:gnn-vg-random-g}

\caption[]{
Qualitative Results on graphs for counterfactuals presented in Fig. 4 of the main paper for VG-RANDOM.}
\label{fig:vg-ran-g}
\end{figure}

In Fig. \ref{fig:vg-dense-g} (VG-DENSE) and \ref{fig:vg-ran-g} (VG-RANDOM) we present the corresponding graphs to counterfactual images of Visual Genome produced by our method and the method of SC \cite{wisely}.

Inspection of VG-DENSE graphs clearly indicates that our method retrieves counterfactual instances that not only have similar concepts on nodes and edges but are also \textbf{structurally closer}. Suggesting counterfactual images with emphasis on object interactions leads to more accurate and meaningful explanations. For instance, in the first column, the relation 'surfer riding board' translates to 'man on board' for our method, whereas for SC \citep{wisely} the  man is essentially holding the board ('cord on board', 'cord on man').

In the case of VG-RANDOM where graphs have many isolated nodes and fewer edges, the comparison is not as straightforward. In columns 1 and 2 of Fig. \ref{fig:vg-ran-g}, our method retrieves visually more similar instances by combining semantics and structure; thus, managing to preserve the main interacting concept of the image. However, when relations are sparse in the source graph, a greater amount of similar concepts will lead to better counterfactuals. 
\newline

% \newpage
\subsection{Actionability of edits}
\label{sec:edits}

We present two non-actionable counterfactual explanations produced by CVE and leverage their method of converting visual CEs into natural language. This approach enables us to precisely define changes between the query and counterfactual instances, which would be challenging with purely visual information. Having emphasized the importance of high-level semantics for human-interpretable CEs, we evaluate the inferred explanations based on linguistic cues rather than pixel-level edits. Both provided examples are deemed successful explanations.

\begin{figure}[t!]
    \centering
    \begin{subfigure}[t]{0.45\textwidth}
        \centering
        \includegraphics[width=6.5cm, height = 3.5cm]{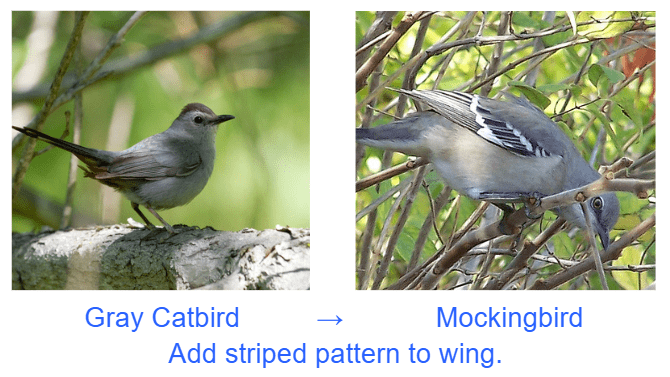}
        \caption{}
        \label{fig:action1}
    \end{subfigure}
     \hspace{0.85cm}
    \begin{subfigure}[t]
    {0.45\textwidth}
        \centering
        \includegraphics[width=6.5cm, height = 3.5cm]{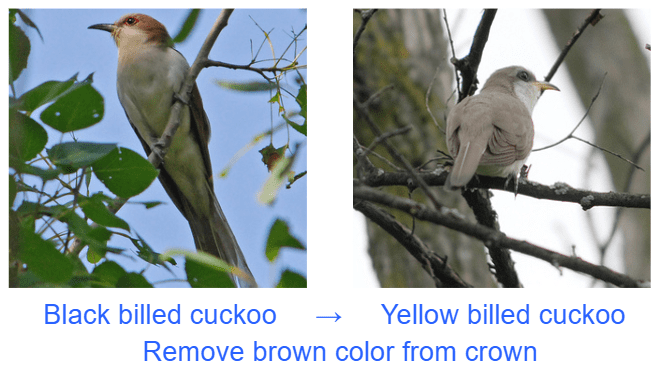}
        \caption{}
        \label{fig:action2}
    \end{subfigure}
    \caption{Counterfactual images from CVE and the proposed explanations using natural language.}
\end{figure}

\cite{vandenhende2022making} often propose single edits on the query image (left images of Figs. \ref{fig:action1}, \ref{fig:action2}) and deem them sufficient for the transition from query to target class.
However, as explained in Sec. 4.1 of the main paper, this approach disregards the rest of the edits needed to be made between $I_{(A)}$ and $I_{(B)}$ and leads to instances that are in fact out-of-distribution. In the main paper, we gave an example that corresponds to Fig. \ref{fig:action1}. In addition to the combination ('has\_head\_pattern::eyering', 'has\_breast\_color::grey') that was reported in text, we provide several other attribute combinations that do not exist in any other bird of the target class in Tab. \ref{tab:pairs}. 

\begin{wraptable}{r}{9cm}
\vskip -0.2in
\caption{Out of distribution attribute pairs for target classes.}
\label{tab:pairs}
\vskip 0.15in
\begin{center}
\begin{tabular}{c}
\toprule
% \hspace{-20px}
\small Gray Catbird $\to$ Mockingbird \\
\midrule
\small ('has\_head\_pattern::eyering', 'has\_breast\_color::grey') \\
\small('has\_head\_pattern::eyering', 'has\_belly\_color::grey')\\
\small('has\_breast\_color::grey', 'has\_nape\_color::brown')\\
\small('has\_breast\_color::grey', 'has\_shape::swallow-like')\\
\small('has\_upper\_tail\_color::white', 'has\_wing\_shape::pointed-wings') \\
\small('has\_breast\_color::grey', 'has\_primary\_color::brown')\\
\small('has\_throat\_color::grey', 'has\_shape::swallow-like')\\
\small('has\_belly\_color::grey', 'has\_shape::swallow-like')\\
\small('has\_shape::swallow-like', 'has\_leg\_color::black')\\

\midrule
\small  Black billed $\to$ Yellow billed Cuckoo \\
 \midrule

\small('has\_upperparts\_color::buff', 'has\_upper\_tail\_color::white') \\
\small('has\_back\_color::white', 'has\_head\_pattern::plain') \\
\small('has\_upper\_tail\_color::white', 'has\_head\_pattern::plain') \\
\small('has\_upper\_tail\_color::white', 'has\_size::very\_small\_(3\_-\_5\_in)') \\
\small('has\_upper\_tail\_color::white', 'has\_back\_pattern::solid') \\
\small('has\_upper\_tail\_color::white', 'has\_leg\_color::buff') \\
\small('has\_head\_pattern::plain', 'has\_nape\_color::white')\\
\small('has\_nape\_color::white', 'has\_back\_pattern::solid') \\
\small('has\_nape\_color::white', 'has\_tail\_pattern::solid') \\
\small('has\_size::very\_small\_(3\_-\_5\_in)', 'has\_bill\_color::grey') \\
\small('has\_leg\_color::buff', 'has\_bill\_color::grey') \\

\bottomrule
\end{tabular}
\end{center}
%\hskip -0.5in
\end{wraptable}
Furthermore, we present one more example in Fig. \ref{fig:action2}. \cite{vandenhende2022making} claim that removal of the brown color from the crown of the Black billed cuckoo in Fig. \ref{fig:action2} (left) is sufficient for it to be classified as a Yellow billed cuckoo.
After performing such an edit we obtain a new bird instance that retains the same features as the bird depicted in Fig. \ref{fig:action2} (left), except it no longer has a brown crown. By generating all pairs of attributes of this new bird, we discover that none of the attribute pairs listed in Tab. \ref{tab:pairs} are representative of any bird in the target class (Yellow billed cuckoo).

It is straightforward to understand that more examples can easily be found throughout the dataset. Given the definition of target classes used in this example (most frequently confused by the classifier),  counterfactual pairs are generally visually and semantically close. If we chose a different definition of the target class and picked one that is dissimilar to the query class, we can deduce that the list of out-of-distribution attribute combinations would be much longer.

% what about our method 
Regarding our method, actionability, in the sense of counterfactuals being representative of the data distribution, is inherent. This guarantee arises from the fact that counterfactuals are actual samples from the target class, specifically the most similar ones to the query, and that we offer complete explanations.To be precise, the proposed counterfactual explanations consist of lists of all graph edits needed to transit from query $I_{(A)}$ to target $I_{(B)}$.

\subsection{Additional results}
\label{sec:additional-res}

\begin{figure}[t]
% \vskip -0.2in
    \centering
    \includegraphics[width=0.57\columnwidth]{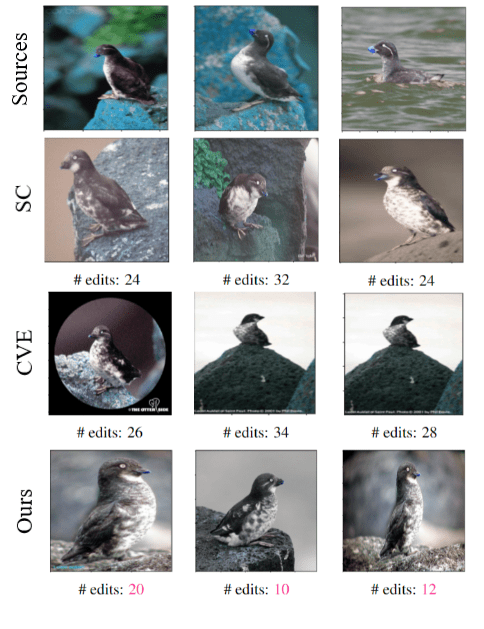}
    \caption{Additional qualitative results of counterfactuals of the source class Parakeet Auklet belonging to target class Least Auklet. We also provide number of total edits per method, with \textcolor{magenta}{colored} instances denoting best results.}
    % \vskip -0.3in
    \label{fig:additional2}
\end{figure}

\paragraph{CUB}
In Fig. \ref{fig:additional2} we provide some additional visual results of counterfactuals comparing our method with SC and CVE. Despite the visual similarity of the retrieved counterfactual images given all three methods, our approach consistently achieves significantly fewer number of total edits.

\section{Applicability to unannotated datasets} 
\label{sec:unannotated}
Applicability to unannotated datasets is a valid concern given our approach's dependence on scene graphs.
As previously established, graphs of images can be obtained either through manual annotations or automated construction methods. However, not all datasets have such readily available resources, therefore we invest our efforts around proving the applicability of our proposed approach to completely unannotated datasets.
% Nevertheless, the issue of applicability arises when considering scenarios where no ground truth annotations are available. 

Studying the impact of annotations is an important aspect, since an intrinsic characteristic of semantic explanations is their dependence on the knowledge of the individual that provides them.
This inherent explainability attribute impacts systems in the same way it does humans. The knowledge supplied to an explainer will determine the specificity and scope of the explanations. Selecting the appropriate annotation technique is a critical step in receiving the desired breadth and depth of explanations.

In the following experiments, we explore these concerns by extracting counterfactual explanations via our proposed framework on unannotated datasets. Our framework is able to explain \textit{any} classifier in a black-box manner, either being a non-neural classifier (humans in the case of the pedestrian vs driver experiment) or a convolution-based model \citep{places} (in the case of Action Genome).

\paragraph{Web images: pedestrian vs driver} \citet{wisely} gather images from Google, Bing, and Yahoo search engines corresponding to 'people', 'motorbikes', and 'bicycles' keywords and their combinations, and then manually split them in 'pedestrian' and 'driver' classes. Finally, 190 'driver' images were obtained (63 images of bicycle drivers and 127 of motorcycle drivers) and 69 'pedestrian' images (31 images of people and parked bicycles, and 38 images of people and parked motorcycles). Those classes are also adopted by us to highlight the importance of relationships (as claimed in \citet{wisely}), as well as extend this claim to support the usage of graphs over the relationship roll-up of SC. By rolling up the roles and converting them into concepts, we might unintentionally overlook important details for a given task. For example, when examining an image depicting a person on a motorbike in a store, alongside another motorbike on the street, by inspecting the scene graph, it is easy to assume that the scene represents a dealership, with the person testing the motorbike for potential purchase, without actually driving it. However, as \cite{wisely} encode this information with the objects: ${person, riding\string.motorbike}$, ${motorbike, in\string.store}$, and ${motorbike, on\string.road}$, they lose the distinction of which motorbike the user is actually riding, potentially leading to erroneous explanations. Nevertheless, leveraging the information within the graph allows us to arrive at more accurate conclusions, especially in fields as critical as Explainable Artificial Intelligence (XAI).

% Employing graphs poses the advantage of 

\begin{figure}[h!]
\centering
\includegraphics[width=0.42\textwidth]{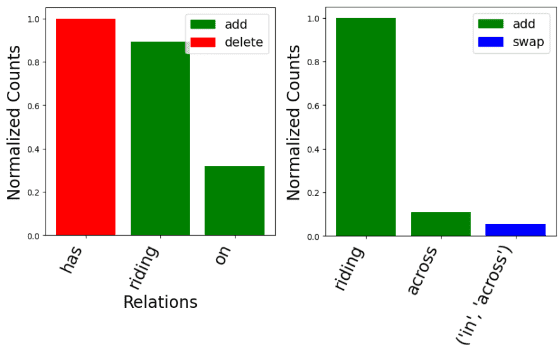}
\caption{Relationships \textcolor{ForestGreen}{inserted}/ \textcolor{red}{deleted}/\textcolor{blue}{substituted} to implement the 'pedestrian' $\rightarrow$ 'driver' transition. }
 \label{fig:relations-sgg}
\end{figure}

Apart from providing triple edits to explain the 'pedestrian' vs 'driver' classification, we also provide global relationship edits to discover if they are meaningful on their own. Indeed, relationship edits are meaningful in general, especially since the 'riding' relationship is inserted frequently (Figure \ref{fig:relations-sgg}, left plot corresponds to immediately deriving the SGG from the image, while the plot on the left denotes the edits occurring from captioning and then obtaining the graph from the caption). Moreover, the relationship 'on' appears frequently (in the SGG case), again confirming the action of sitting on a bike/motorcycle in order to drive.

\begin{figure}[h!]
\centering
\includegraphics[width=0.42\textwidth]{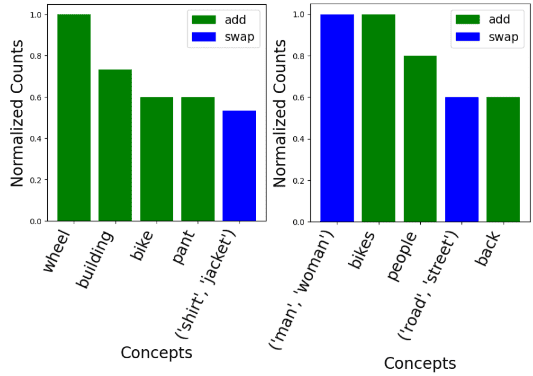}
\caption{Concepts \textcolor{ForestGreen}{inserted}/ \textcolor{red}{deleted}/\textcolor{blue}{substituted} to implement the 'pedestrian' $\rightarrow$ 'driver' transition. }
 \label{fig:concept-sgg}
\end{figure}

Similarly, we extract global edits for concepts discriminating the pedestrian/driver categories. These edits are presented in Figure \ref{fig:concept-sgg}. By observing these plots (SGG - left, captioning and graph from text - right) we conclude that these edits are not really meaningful according to human perception: inserting wheels does not explain the 'pedestrian' $\rightarrow$ 'driver' transition, since in both classes bike/motorbike wheels may appear as part of these vehicles. The same observation is valid for the rest of the concepts appearing on these plots, resulting in noisy conceptual edits. 
To this end, we verify that \textit{explanations are human-dependable}, i.e. a human is the final evaluator of any explanation, and while a method is able to provide semantically meaningful explanations (in this case relationship edits), it is possible that at the same time the same method provides meaningless explanations (in this case concept edits). Nevertheless, if the derived explanations are not conceptual, a human cannot verify their validity; therefore, we can safely claim that \textit{human interpretability of explanations is highly tied to semantics}.

\begin{figure}[h]
    \centering
    \begin{subfigure}{0.7\textwidth}
        \centering
        \includegraphics[width=0.8\linewidth, height=6.5cm]{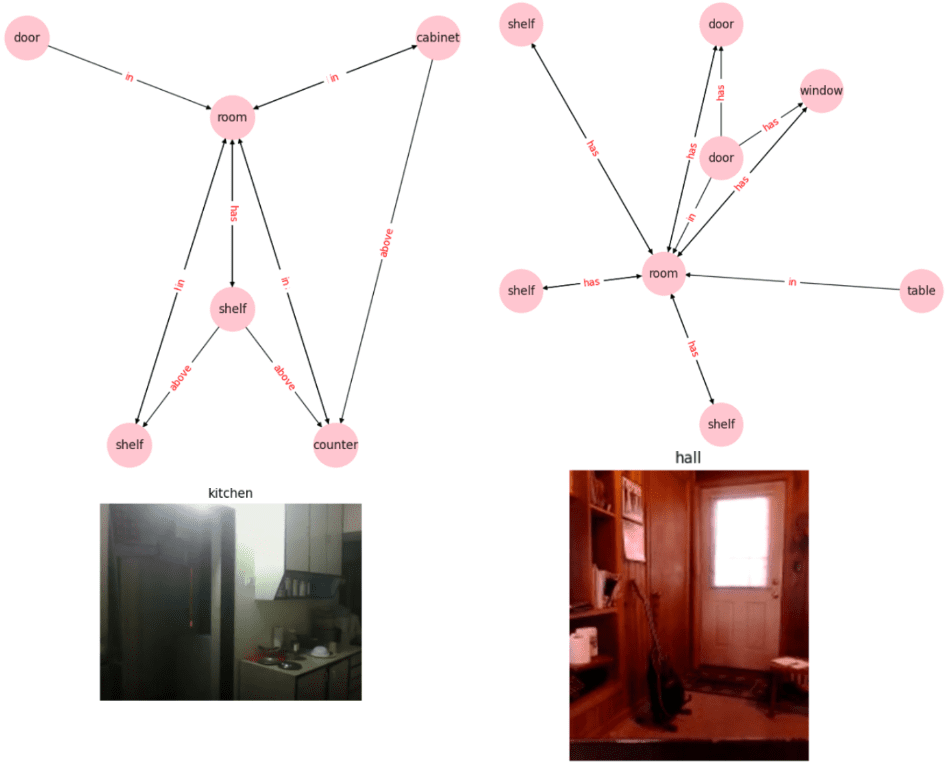}
        \caption{}
        \label{fig:ag-qual1}
    \end{subfigure}
     ~
    \begin{subfigure}{0.7\textwidth}
        \centering
        \includegraphics[width=0.8\linewidth, height=6.5cm]{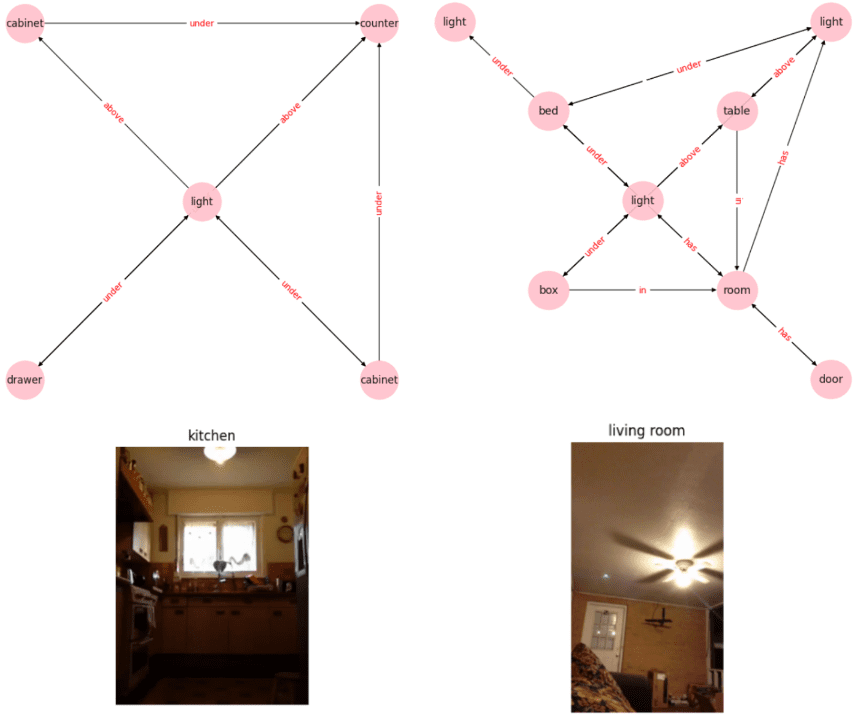}
        \caption{}
        \label{fig:ag-qual2}
    \end{subfigure}    
    ~
    \begin{subfigure}{0.7\textwidth}
        \centering
        \includegraphics[width=0.8\linewidth, height=5.5cm]{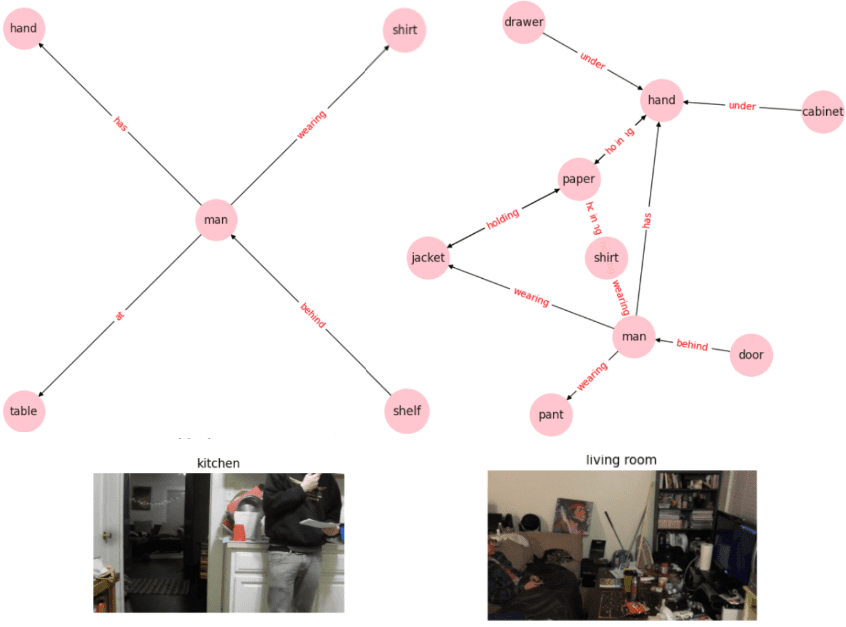}
        \caption{}
        \label{fig:ag-qual3}
    \end{subfigure}
    \caption{Counterfactual examples from AG dataset for query images belonging to the class "kitchen". Here, CEs are classified as "hall" or "living room".}
    \label{fig:ag-qual}
\end{figure}

% \newpage

\paragraph{Action Genome}
We test our method in a real-world image dataset extracted from Action Genome \citep{ji2020action},
% consisting of individual frames from videos. 
a video database depicting human-object relations and actions. It is completely unannotated and also like VG has no predetermined classes for its instances. 
AG results are not presented in the main paper because they offer no new insights compared to other extendability experiments. However, a brief qualitative analysis was deemed interesting enough to present in the appendix.
We select a subset of 300 individual frames and generate scene graphs following well-established SGG methods\footnote{\href{https://medium.com/stanford-cs224w/scene-graph-generation-compression-and-classification-on-action-genome-dataset-9f692a1d5394}{SGG on Action Genome}}. 
After applying our CE method using predictions made by \cite{places}, we obtain results comparable to previous experiments.
% that utilized ground truth scene graphs. 
Specifically, the binary retrieval metrics ranged from 0.17 - 0.41 for P@k and 0.21 - 0.35 for NDCG@k, while overall average edits were 12.86. 
This experiment validates the relative ease of obtaining graphs from images and demonstrates the applicability of our method to AI-generated graphs of varying quality.

% \paragraph{Action Genome}
% Qualitative results for Action Genome (AG) accompanied by their scene graphs are provided in Fig. \ref{fig:ag-qual}.  

% We additionally tested our method on the Action Genome (AG) dataset that 

In Fig. \ref{fig:ag-qual}, we offer some qualitative results on the AG dataset. Instances in this custom AG subset are individual video frames that depict mostly indoor spaces with or without people at a variety of angles and settings. Due to the lack of control in this case, we have identified specific categories that are more meaningful to human perception, such as 'kitchen', 'hall',  and 'living room'. 

By observing these examples, we can initially note that automatically generated graphs provide a satisfactory representation of the images. However, there are missing details and known biases resulting from imbalanced triple and relation distributions in VG, where the SGG models are trained. We analyze the counterfactuals while acknowledging the potentially lower quality of the input graphs. Since this part of the experiments aims to demonstrate the applicability of our method to unannotated datasets, in-depth analysis is not performed. Nonetheless, we can observe that the retrieved graphs exhibit structural similarities and share common concepts, which is also visually apparent. For instance, images featuring kitchens often involve the removal of cabinets located above counters, while tables are prevalent in hallway depictions.

\section{Applicability on other modalities}
\label{sec:smarty}
\begin{table}[h!]
\centering
\caption{Global triple and concept edits for COVID-19 Negative $\rightarrow$ Positive.}
\label{tab:n_edits-cub}
\label{table:covid-appendix}
\vskip 0.15in
\begin{tabular}{cc}
\toprule
\small Concept Edits& \small Normalized Counts  \\ \midrule
\small 'Sneezing' & \small 1.0 \\\small 'RunnyNose' & \small 0.78 \\
\small 'DryThroat' & \small 0.35 \\
\small 'Fever'& \small 0.34 \\
\small  'Dizziness' & \small 0.31 \\
 \small  'Fatigue'&\small 0.22\\
 \small  'Respiratory'&\small 0.22\\
 \small  'DryCough'&\small 0.21\\
 \small  'TasteLoss'&\small 0.21\\
 \small  'Cough'&\small 0.16\\
 \midrule
\small Triple Edits& \small Normalized Counts  \\ \midrule
\small 'Sneezing' & \small 1.0 \\\small 'RunnyNose' & \small 0.73\\
\small ('Male', 'Female')& \small 0.68\\
 \small 'DryThroat'&\small 0.36\\
\small 'Fever'& \small 0.35\\
\small  'Dizziness' & \small 0.31 \\
 \small ('Fourties', 'Twenties')&\small 0.29\\
 \small  'DryCough'&\small 0.23\\
 \small  'Fatigue'&\small 0.23\\
 \small  'Respiratory'&\small 0.23\\
 \midrule
\end{tabular}
\end{table}
We will provide some details on the modality-agnostic nature of our approach, and specifically results on audio classification.

The process of SMARTY graph generation differs compared to our previous experiments in a few key ways. In this new approach, each user or patient was directly connected to their symptoms and characteristics, which were defined to be audible to a certain extent. Symptom analysis involved treating certain symptoms as sub-symptoms when necessary, based on the hierarchical structure presented in \citet{wisely}'s SMARTY hierarchy, as opposed to using WordNet for computing node edit costs. Regarding edges in the graph, a simpler strategy was adopted due to the limited number of edge types. Specifically, the approach considered edge swaps between different edge types, as well as the addition and deletion of edges, as costly operations. 

To initialize the GNN similarity component, custom BioBert \citep{lee2020biobert} embeddings were utilized because the language used in the medical field is specific and distinct from general language, unlike previous approaches that relied on simple Glove embeddings. These changes were made to enhance the accuracy and relevance of the SMARTY graph generation.

In Table \ref{table:covid-appendix} comprehensive global edit lists can be found.  It is important to note that in Table \ref{table:covid-appendix}, triple edits refer to edge edits and the concepts adjacent to them. For the sake of readability, we have omitted the head and predicate of the triples, where all heads are the 'User' concept and all predicates represent symptoms or sub-symptoms. The second half of  \ref{table:covid-appendix} on the other hand, focuses on node edits, regardless of edges. Evidently, there is agreement with Table \ref{table:covid-appendix}, but there are also additional noteworthy findings. One of these findings relates to the reported gender bias mentioned in \citet{wisely}, and another suggests a correlation between COVID-19 positivity and younger users.

\section{Limitations}
\label{sec:limitations}
Our work is subject to certain limitations. 
First of all, our experiments involving the CUB and VG datasets are highly dependent on the existing annotations, thus influencing the quality of the derived conceptual explanations. Specifically, the generated semantics through SGG are influenced by the training datasets, namely VG. This limitation was addressed through the comparison of our method's consistency among two vastly different graph generation methods. Despite the positive results validated by the similar produced global edits, there is much room for exploration in this domain. We plan to engage in this venture in our future research. 
Moreover, pre-trained image classifiers, such as ResNet50 and Places365 may produce imperfect labels for the images under consideration, which may influence the resulting counterfactual explanations.
CEs are also characterized by known limitations, such as robustness \citep{slack2021counterfactual}. While we have not addressed this particular limitation in our work, we plan to explore it in our future work. Despite these limitations, we have ensured actionability guarantees with the aim of improving the quality of the provided counterfactuals.

%% file: tables-app/cub_avg_ged.tex
\begin{wraptable}{r}{4cm}
    \vskip -0.2in
    \centering
        \caption{Average top-1 GED on CUB. \textbf{Bold numbers} denote best results.
    }
    % \vskip -0.04in
    \label{tab:avg-ged-cub}
    \vskip 0.15in
    \begin{tabular}{c|P{1.7cm}} 
        \toprule
        & CUB $\downarrow$\\

        \hline
        \small CVE  & \small 257.20  \\
        \small SC  & \small 263.80  \\
        \small Ours & \small \textbf{211.69} \\
        \bottomrule
    \end{tabular} 
    % \vskip -0.09in
\end{wraptable}

%% file: tables-app/refined_num_edits.tex
\begin{table}[h!]
% \vskip -0.17in
\centering
\caption{Refined average number of node, edge \& total edits on VG.
% \textbf{Bold} denotes best results.
}
\vskip 0.14in
\label{tab:n_edits-vg-refined}
\begin{tabular}{c|ccc|ccc} 
\toprule
&  \multicolumn{3}{c}{\small VG-DENSE} &  \multicolumn{3}{c}{\small VG-RANDOM} \\
\cline{2-7}
& \small Node$\downarrow$ & \small Edge$\downarrow$ & \small Total$\downarrow$& \small Node$\downarrow$ & \small Edge$\downarrow$ & \small Total$\downarrow$\\
\hline
\small SC  &  \small \textbf{4.73}&  \small 7.65&  \small 12.38&  \small \textbf{11.96}&  \small \textbf{7.48}&  \small \textbf{19.44}\\
\small Ours &  \small 5.07&  \small \textbf{6.96}&  \small \textbf{12.03}&  \small 12.37&  \small 7.52&  \small 19.89\\
\bottomrule
\end{tabular} 
% \vskip -0.09in
\end{table}